\documentclass{article}

\PassOptionsToPackage{authoryear}{natbib}

\usepackage[dblblindworkshop, final]{neurips_2026}
\workshoptitle{Neural Network Artifacts as a New Data Modality}

\usepackage[utf8]{inputenc} %
\usepackage[T1]{fontenc}    %
\usepackage{hyperref}       %
\usepackage{url}            %
\usepackage{booktabs}       %
\usepackage{amsfonts}       %
\usepackage{nicefrac}       %
\usepackage{microtype}      %
\usepackage{xcolor}         %
\usepackage{graphicx}
\usepackage{subcaption}
\usepackage{wrapfig}
\usepackage{tcolorbox}
\tcbuselibrary{listings}

\newtcblisting{promptbox}{
    colback=gray!15,
    colframe=gray!50,
    boxrule=0.7pt,
    arc=2pt,
    listing only,
    listing options={
        basicstyle=\ttfamily\small,
        breaklines=true,
        columns=fullflexible,
    },
}

\title{Where Does Alignment Emerge and \\ What Does It Mean? A Closer Look at the \\ Platonic Representation Hypothesis}

\author{
Julia Herbst$^{1,2}$ \quad Simon Roschmann$^{1,2,3}$ \quad Quentin Bouniot$^{1,2,3,4}$ \\ \textbf{Zeynep Akata}$^{1,2,3}$\\
\\
$^1$Helmholtz Munich \quad
$^2$Technical University of Munich \\
$^3$Munich Center for Machine Learning \\
$^4$LTCI, Télécom Paris, Institut Polytechnique de Paris, Palaiseau, France\\
\\
\texttt{\{julia.herbst,simon.roschmann,zeynep.akata\}@tum.de} \\ 
\texttt{quentin.bouniot@telecom-paris.fr}
}

\begin{document}

\maketitle

\begin{abstract}
The Platonic Representation Hypothesis (PRH) suggests that neural networks are converging toward similar representation spaces. As language models become more capable, their representations are aligning with those of vision models. However, recent work has questioned whether this trend continues to hold for newer language models. Moreover, it has been shown that representation similarity metrics can be confounded by model depth and width, requiring appropriate baselines for interpretation. In this work, we first investigate where alignment emerges between unimodal models across model layers and representation dimensions. Next, we find that the PRH capability-alignment trend persists for newer models as long as the captions are sufficiently clean and informative. Finally, we contextualize cross-modal alignment, introducing unimodal and multimodal baselines on various datasets. Our results show that the cross-modal representational alignment of unimodal models can surpass that of explicitly aligned vision-language models, but remains substantially behind the alignment observed for models of the same modality. Code is available at \href{https://github.com/ExplainableML/prh-closer-look}{\texttt{github.com/ExplainableML/prh-closer-look}}.
\end{abstract}

\section{Introduction}

Images and textual descriptions provide two views of the same underlying world. The Platonic Representation Hypothesis (PRH) \citep{huh2024platonic} posits that sufficiently capable vision and language models may therefore learn similar abstractions despite being trained on different modalities. This insight has motivated a series of methods \citep{huang2026llm2clip,maniparambil2025freezealign,roschmann2026sotalign,vouitsis2024fusemix,yacobi2025sue,zhang2025sail} that align pretrained unimodal models with lightweight alignment layers and limited paired data instead of training for cross-modal alignment from scratch \citep{radford2021clip,tschannen2025siglip2,zhai2023siglip}.

Recent work has refined the PRH. \citet{koepke2026back} show that some findings strongly depend on the evaluation protocol and that performance-alignment trends observed for earlier language models no longer extend to newer model families. \citet{groger2026revisiting} argue that similarity scores are difficult to interpret without suitable baselines, and thus propose calibrating them against a null hypothesis representing the absence of alignment. Moreover, these investigations into the PRH commonly report the maximum similarity across all layer pairs using the full representation of models, obscuring at which layers and dimensions alignment emerges.

We take a closer look at the PRH along three axes. First, we analyze where alignment emerges across model layers and dimensions under compression via principal component analysis (PCA). Second, we examine how the relationship between language model capability and cross-modal alignment is affected by caption characteristics. Third, we contextualize the cross-modal alignment of unimodal vision and language models beyond a null baseline. To our knowledge, no prior work has provided a comparison with randomly initialized models, explicitly aligned vision-language models (VLMs), and model pairs from the same modality.

\section{Experimental setup}
\label{sec:experimental-setup}

\paragraph{Models.}
We evaluate a broad set of unimodal models. On the vision side, we consider DINOv2 ViT-L~\citep{oquab2023dinov2}, DINOv3 ViT-L~\citep{simeoni2025dinov3} and MAE ViT-L~\citep{he2021maskedautoencodersscalablevision}. 
On the language side, we evaluate 21 generative models from the
BLOOMZ~\citep{muennighoff2022crosslingual},
Llama~\citep{touvron2023llama,grattafiori2024llama, meta2024llama32},
Mistral~\citep{jiang2023mistral7b},
Qwen~\citep{qwen2,qwen2.5,qwen3technicalreport},
Gemma~\citep{gemmateam2024gemmaopenmodelsbased,gemma_2024,gemmateam2024gemma2improvingopen}, and
DeepSeek~\citep{deepseekai2025deepseekr1incentivizingreasoningcapability}
families. Additionally, we include the text embedding models
Qwen3-Embedding-8B~\citep{zhang2025qwen3embedding} and
NV-Embed-v2~\citep{lee2025nvembed}.
We compare their alignment against explicitly aligned VLMs such as CLIP ViT-L/14~\citep{radford2021clip,ilharco2021openclip,schuhmann2022laion}, SigLIP 2 So400M~\citep{tschannen2025siglip2} and Qwen3-VL-Embedding-8B~\citep{qwen3vlembedding}. For the explicitly aligned VLMs, we extract image and text representations from the joint embedding space after the final projection layer. The text encoders of CLIP and SigLIP~2 support a maximum context length of 77 and 64 tokens, respectively. For captions exceeding these limits, we split the text into chunks that respect the context window, embed each chunk independently, and finally average these chunk representations.

\paragraph{Datasets.}
We evaluate alignment on image-text datasets with different caption properties: WIT~\citep{srinivasan2021wit} with short and noisy captions from Wikipedia, CC3M~\citep{sharma2018conceptual} with short and synthetically created DreamLIP captions~\citep{zheng2024dreamlip}, COCO~\citep{lin2014microsoft} with concise human-written captions, and DOCCI~\citep{onoe2024docci} with long and fine-grained human-written captions. To investigate the effect of caption verbosity for the same set of images, we additionally construct shortened versions of the DOCCI captions. Specifically, we prompt Qwen3-8B \citep{qwen3technicalreport} to summarize each caption to approximately 10\%, 25\%, 50\% or 75\% of its original length while preserving the most salient semantic content. The prompt is provided in Appendix~\ref{apd:prompt} along with a DOCCI sample and its newly generated captions.

\paragraph{Metric.}
We measure alignment using mutual $k$-nearest-neighbor overlap \citep{huh2024platonic} and its calibrated variant proposed by \citet{groger2026revisiting}. For each dataset, we sample $N=1024$ image-caption pairs and compare the indices of the $k=10$ nearest images in vision space with those of the nearest captions in language space. \citet{koepke2026back} find that alignment decreases for fixed $k$ with increasing gallery size, but also show that it remains stable when the relative neighborhood size $k/N$ is preserved. An ablation study on $k$ and $N$ is provided in Appendix \ref{app:mutualkNN}. We use the CLS token for vision models and mean-pooled representations for language models and apply $L_2$ normalization.

\paragraph{Implementation.} All experiments can be performed on a single NVIDIA A100 GPU with 80 GB memory. We use the LM Evaluation Harness framework~\citep{eval-harness} to obtain the MMLU and ARC Challenge accuracies as well as the GSM8K exact match scores for the generative language models in Section~\ref{subsec:performance-alignment}.

\begin{figure*}[t]
\centering
\begin{subfigure}{0.32\textwidth}
\centering
\includegraphics[width=\linewidth]{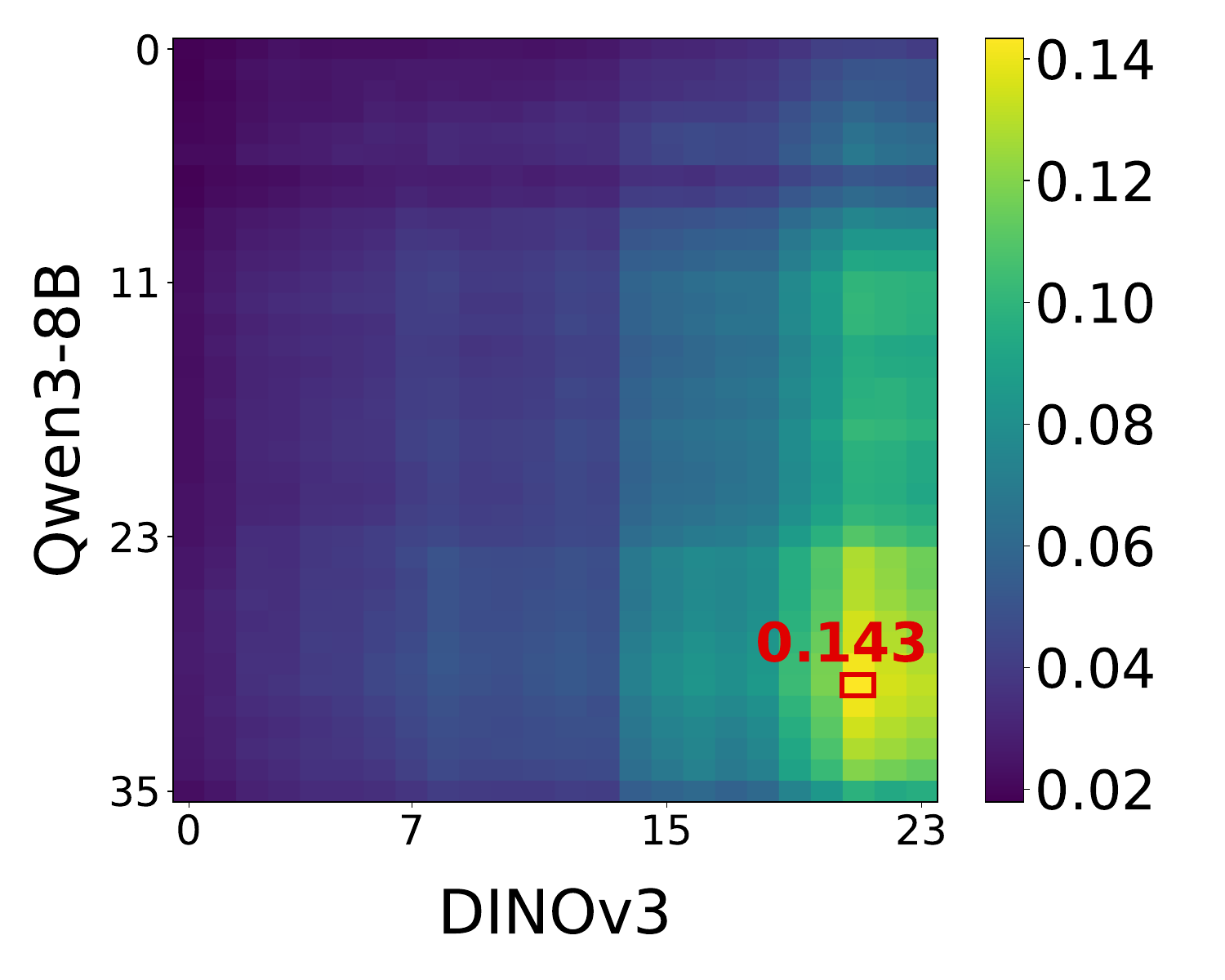}
\caption{Vision-Language}
\label{fig:heatmap_alignment_layers_a}
\end{subfigure}
\hfill
\begin{subfigure}{0.32\textwidth}
\centering
\includegraphics[width=\linewidth]{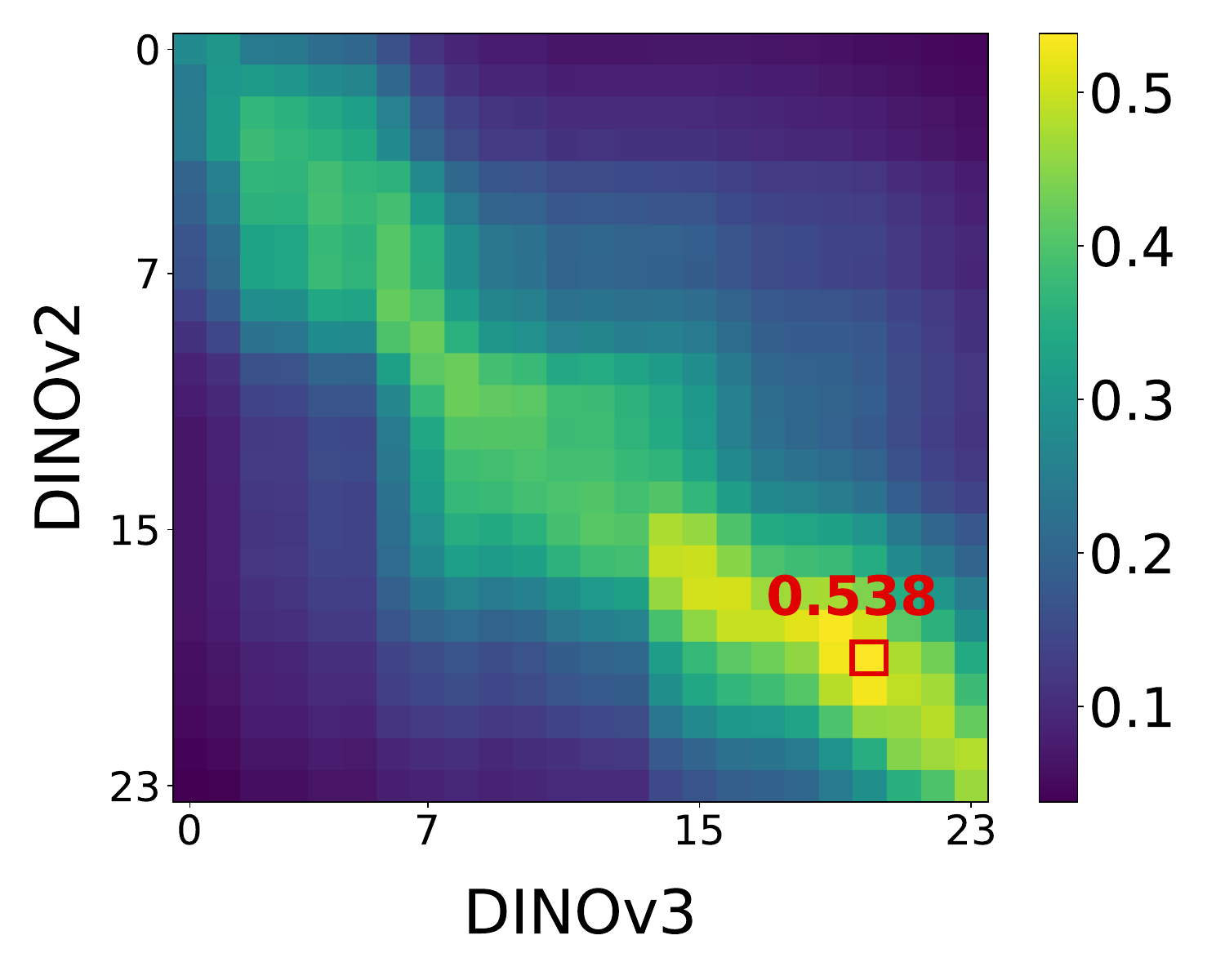}
\caption{Vision-Vision}
\label{fig:heatmap_alignment_layers_b}
\end{subfigure}
\hfill
\begin{subfigure}{0.32\textwidth}
\centering
\includegraphics[width=\linewidth]{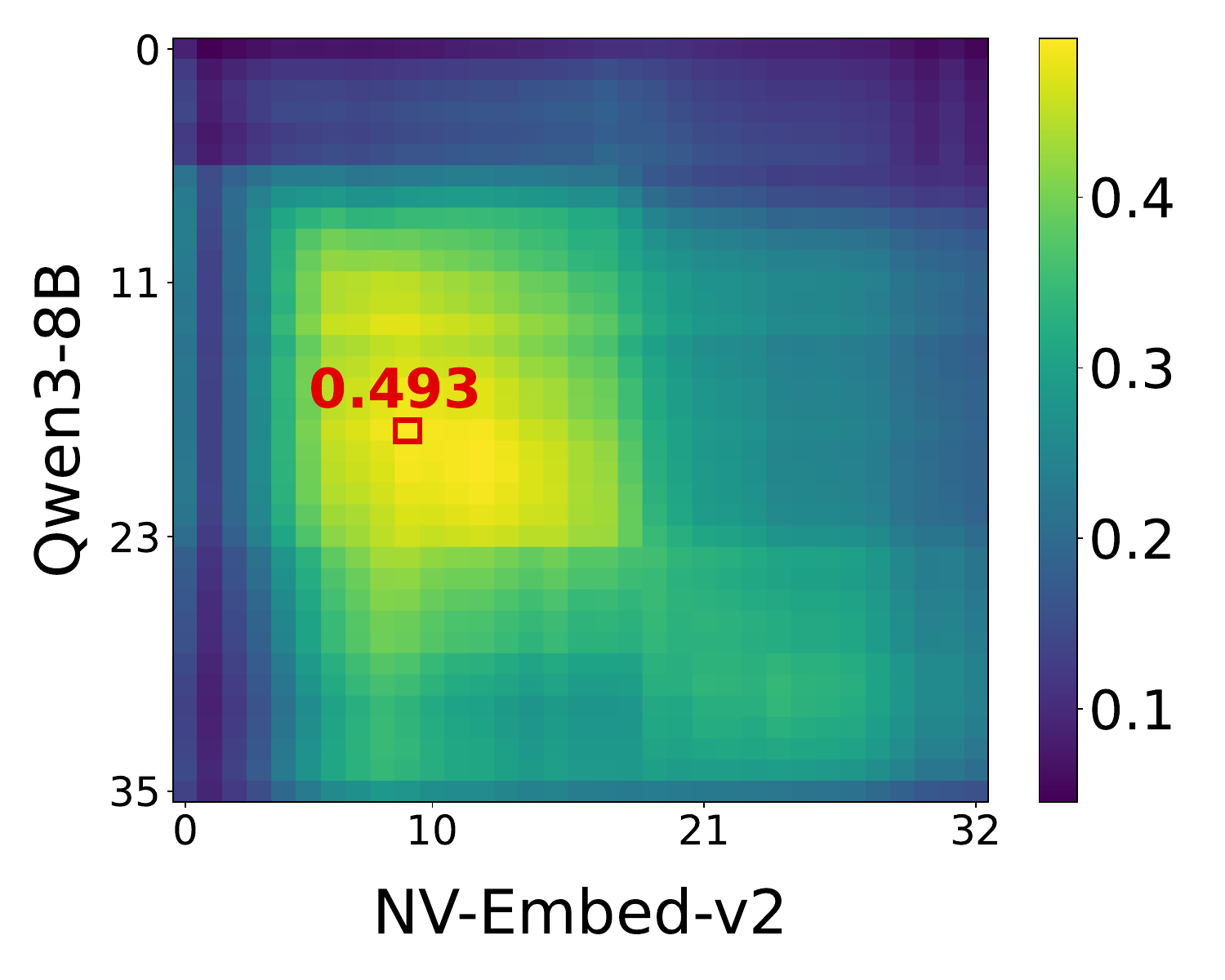}
\caption{Language-Language}
\label{fig:heatmap_alignment_layers_c}
\end{subfigure}
\caption{Heatmaps with mutual $k$-NN alignment between every layer on WIT.}
\label{fig:heatmap_alignment_layers}
\end{figure*}

\section{Evaluation}

\subsection{Where does alignment emerge across layers and dimensions?}
\label{subsec:where}

Prior work \citep{huh2024platonic, koepke2026back,groger2026revisiting} reports only the maximum alignment across layer pairs. Figure~\ref{fig:heatmap_alignment_layers} instead shows the complete layer-wise alignment between two models on WIT. Cross-modal alignment emerges mainly in later layers, once more abstract semantic representations have formed, and peaks at a mutual $k$-NN score of $0.143$ between DINOv3 layer 21 and Qwen3-8B layer 30. Within-modality alignment is much stronger. Vision models form three broad block-diagonal regions above $0.3$, peaking at $0.538$ among later layers. A generative language model (Qwen3-8B) and a language embedding model (NV-Embed-v2) reach a mutual $k$-NN score around $0.5$ across many early-to-middle layer pairs. Notably, final-layer representations never reach maximum alignment and may even be far from it. More layer-wise alignment results are presented in Appendix~\ref{apd:layer-wise_analysis}.%

\begin{figure}[t]
    \centering
    \begin{subfigure}[t]{0.48\linewidth}
        \centering
        \includegraphics[width=\linewidth]{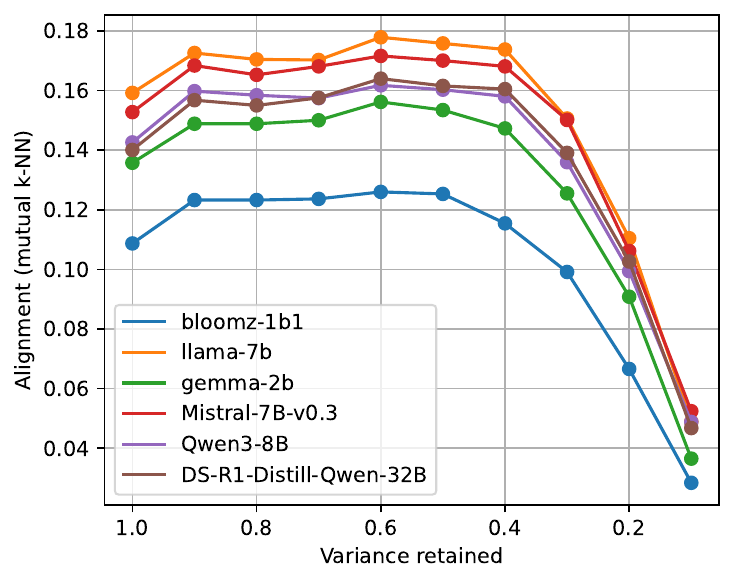}
        \caption{Explained variance.}
        \label{fig:compression_var}
    \end{subfigure}
    \hfill
    \begin{subfigure}[t]{0.48\linewidth}
        \centering
        \includegraphics[width=\linewidth]{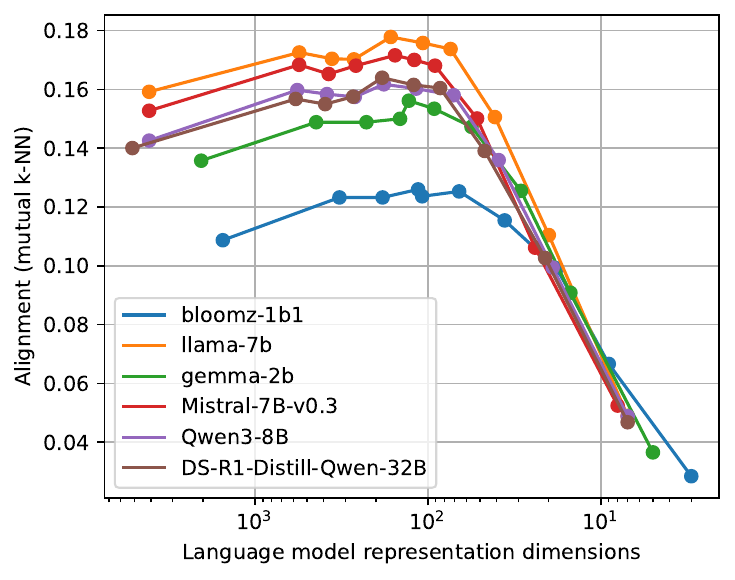}
        \caption{Number of dimensions.}
        \label{fig:compression_dims}
    \end{subfigure}
    \caption{Effect of PCA compression on alignment between DINOv2 and language models on WIT.}
    \label{fig:compression}
\end{figure}

We next investigate whether alignment is localized not only to specific layers, but also within a small subspace of the representations. Specifically, we compress the vision and language representations via PCA before computing mutual $k$-NN on WIT. PCA is fit independently for each layer of each model, with separate transformations for the language and vision representations. Figure~\ref{fig:compression} shows that retaining 40\% of the variance preserves or slightly improves alignment for most models, suggesting that lower-variance directions mainly capture modality-specific or noisy information. However, very strong compression causes alignment to collapse, indicating that the shared structure remains distributed across multiple principal components. In Figure~\ref{fig:compression_dims}, we plot alignment against the number of dimensions retained after PCA. This allows us to investigate the effective dimensionality of the subspaces in which the observed alignment resides. In particular, the results suggest that the cross-modal alignment is largely captured within a relatively small subspace of approximately 100 dimensions, compared to the original language spaces with 1,536--5,120 dimensions. Additional results on DOCCI and COCO are provided in Appendix~\ref{apd:compression_pca}.

Together, these results indicate that cross-modal alignment is localized both across depth and within the representation space: only particular layers exhibit strong correspondence, and within those layers, the correspondence can be captured by relatively low-dimensional subspaces.

\begin{figure*}[ht]
\centering
\begin{subfigure}{0.48\linewidth}
\centering
\includegraphics[width=\linewidth]{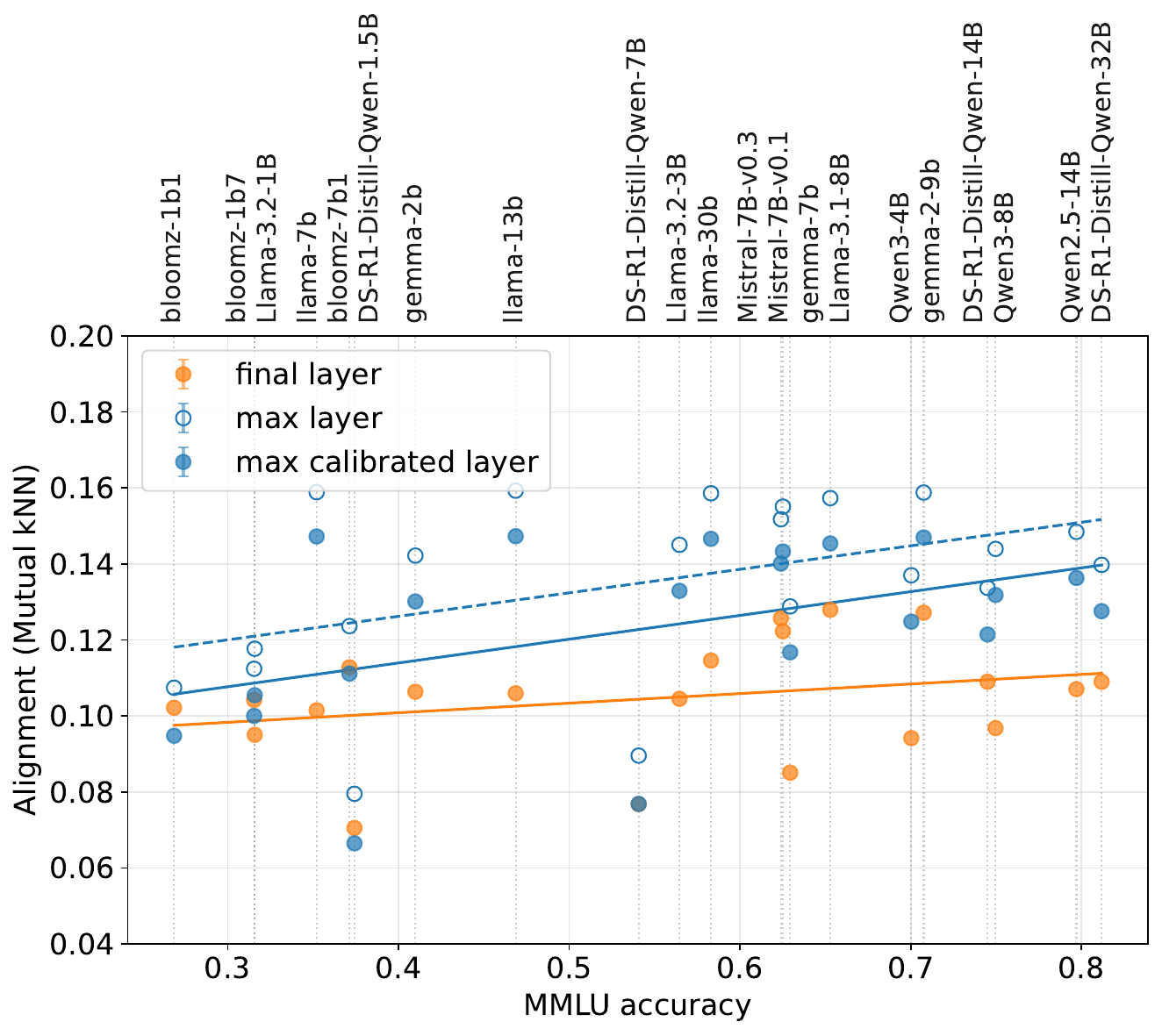}
\caption{WIT.}
\label{fig:mmlu_alignment_wit_app}
\end{subfigure}
\hfill
\begin{subfigure}{0.48\linewidth}
\centering
\includegraphics[width=\linewidth]{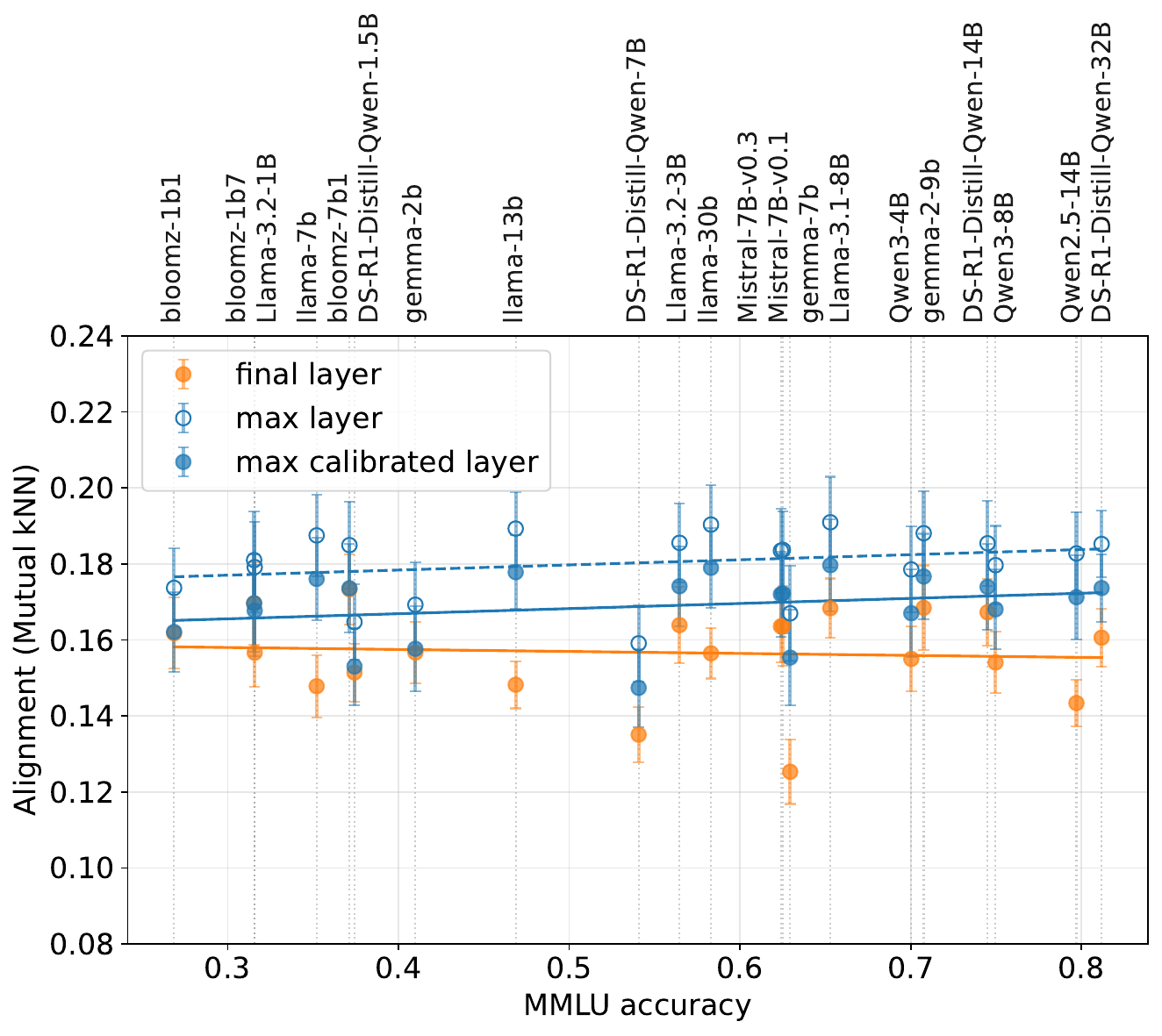}
\caption{CC3M short synthetic.}
\label{fig:mmlu_alignment_cc3m_short_app}
\end{subfigure}
\begin{subfigure}{0.48\linewidth}
\centering
\includegraphics[width=\linewidth]{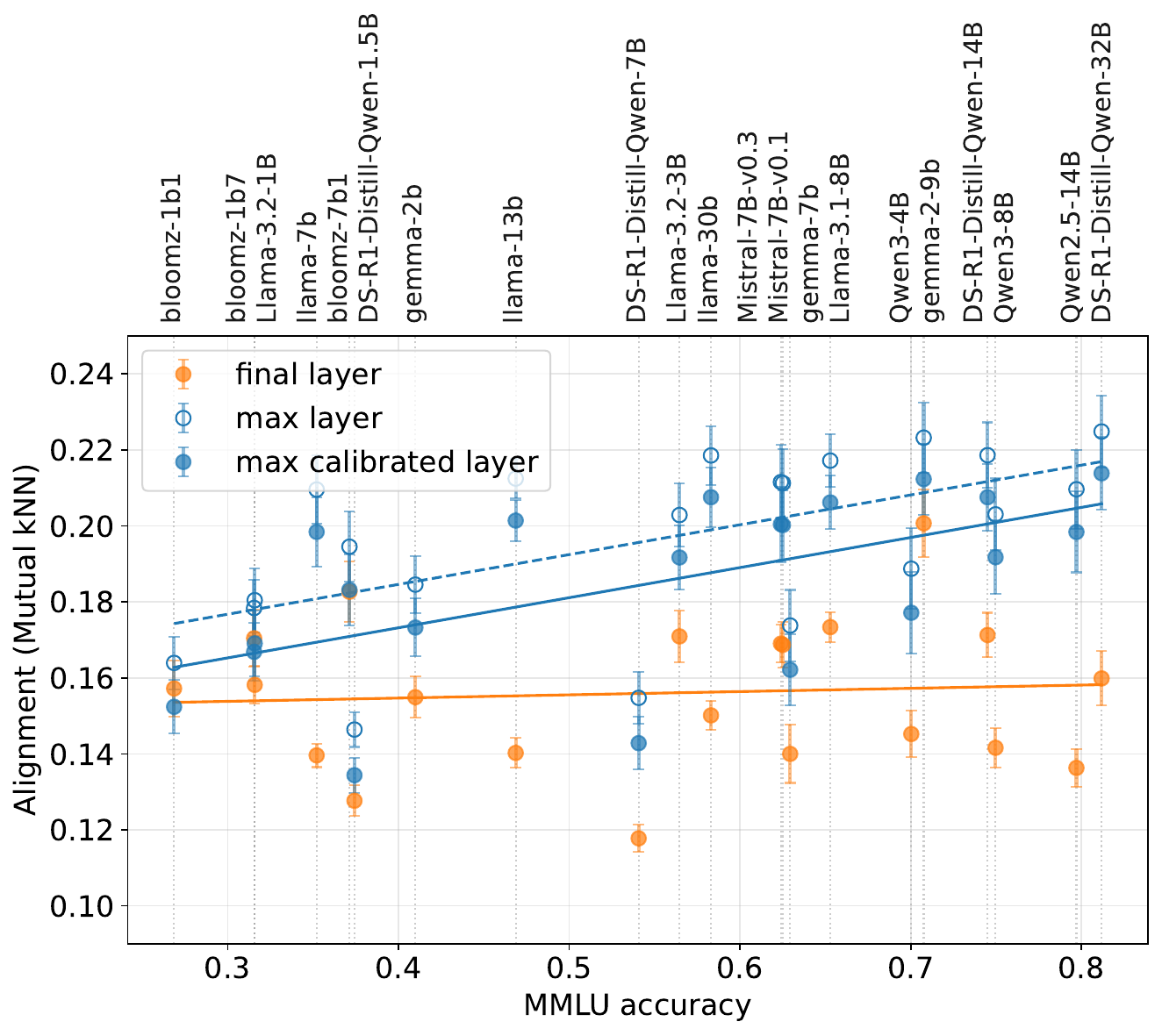}
\caption{DOCCI short synthetic.}
\label{fig:mmlu_alignment_docci_short_app}
\end{subfigure}
\hfill
\begin{subfigure}{0.48\linewidth}
\centering
\includegraphics[width=\linewidth]{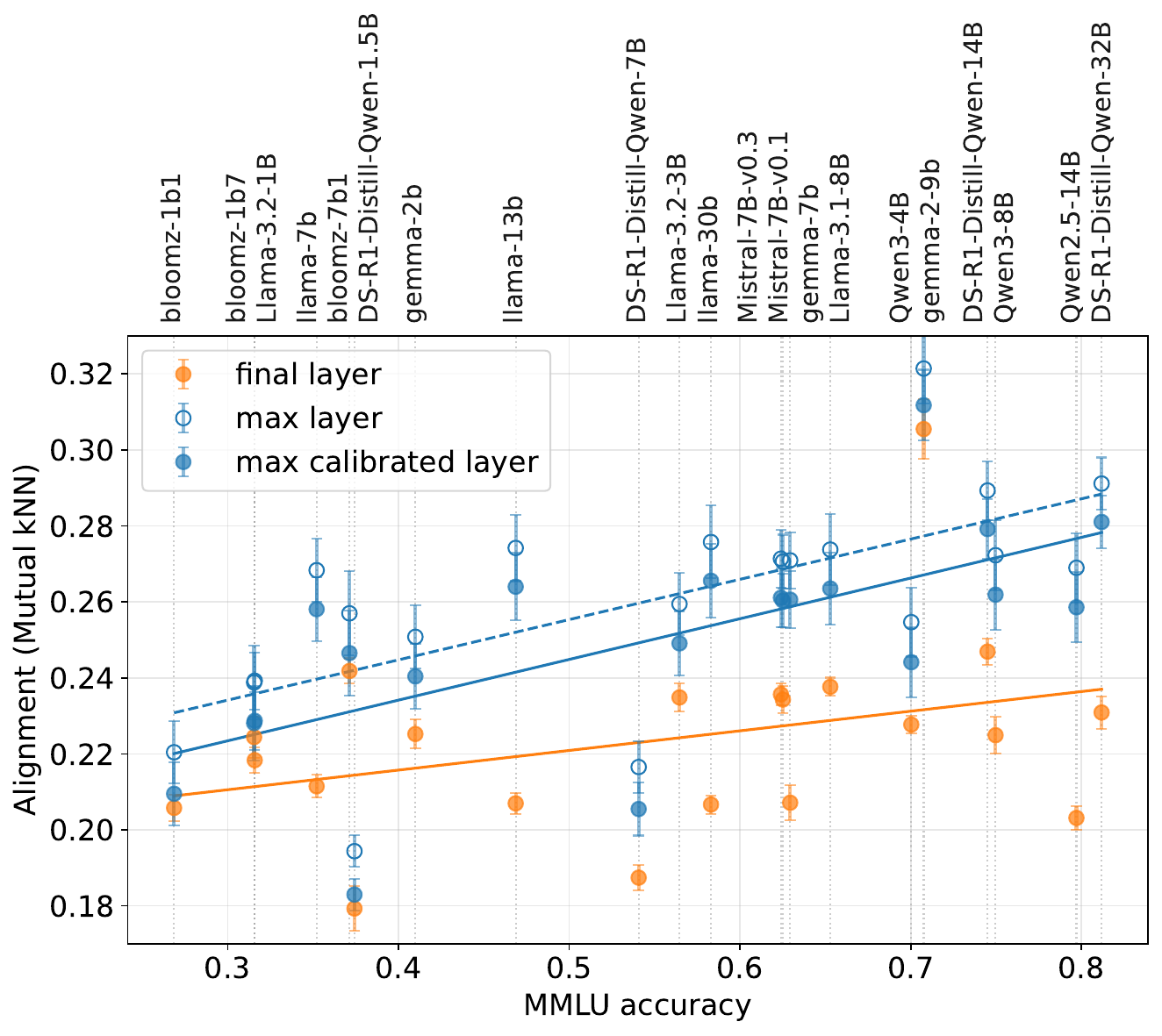}
\caption{DOCCI.}
\label{fig:mmlu_alignment_docci_long_app}
\end{subfigure}
\caption{Performance of language models on MMLU vs.\ alignment with vision models.}
\label{fig:mmlu_alignment_four_datasets}
\end{figure*}

\begin{figure*}[ht]
\centering
\begin{subfigure}{0.48\linewidth}
\centering
\includegraphics[width=\linewidth]{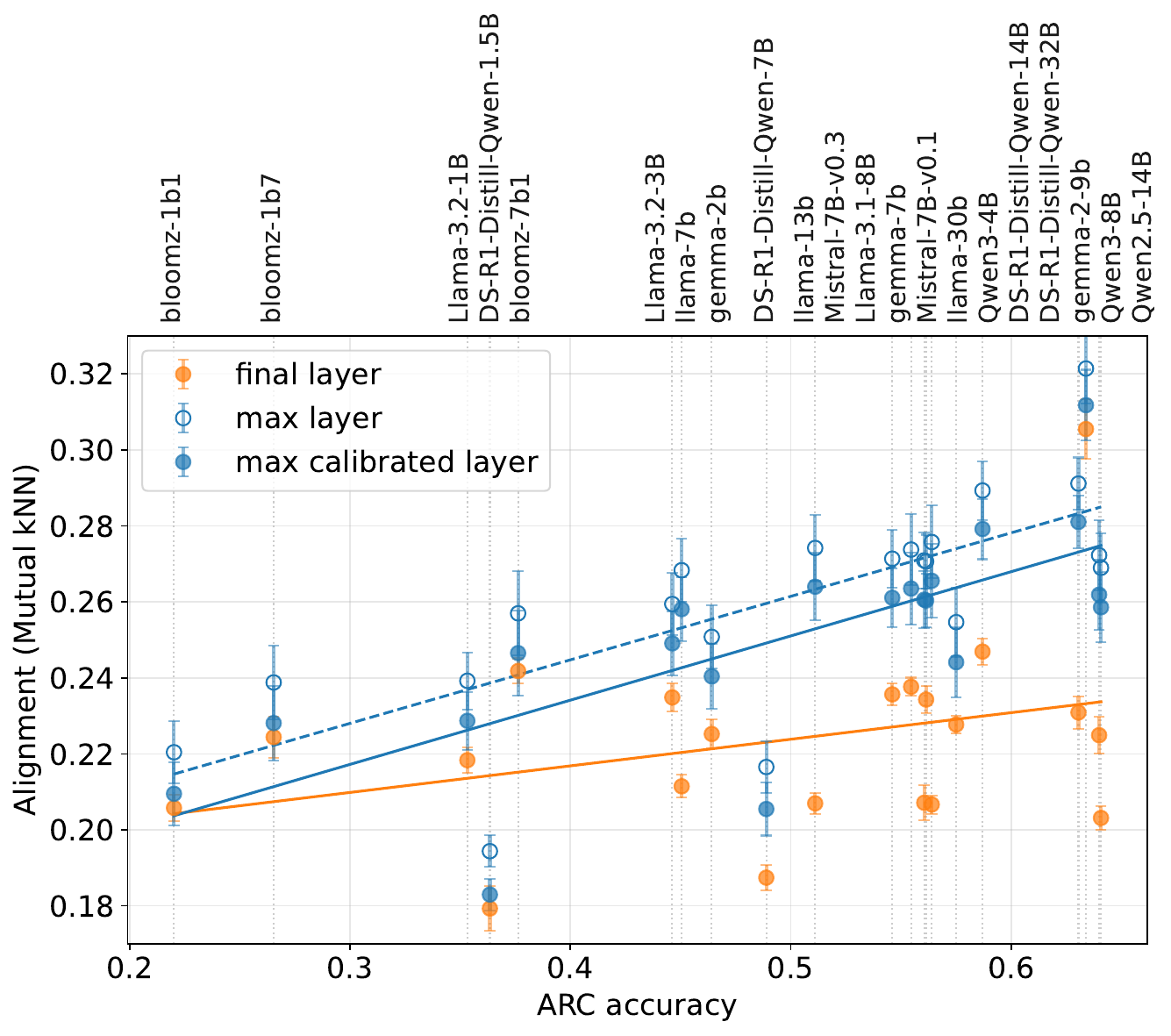}
\caption{ARC accuracy.}
\label{fig:arc_alignment_docci_app}
\end{subfigure}
\hfill
\begin{subfigure}{0.48\linewidth}
\centering
\includegraphics[width=\linewidth]{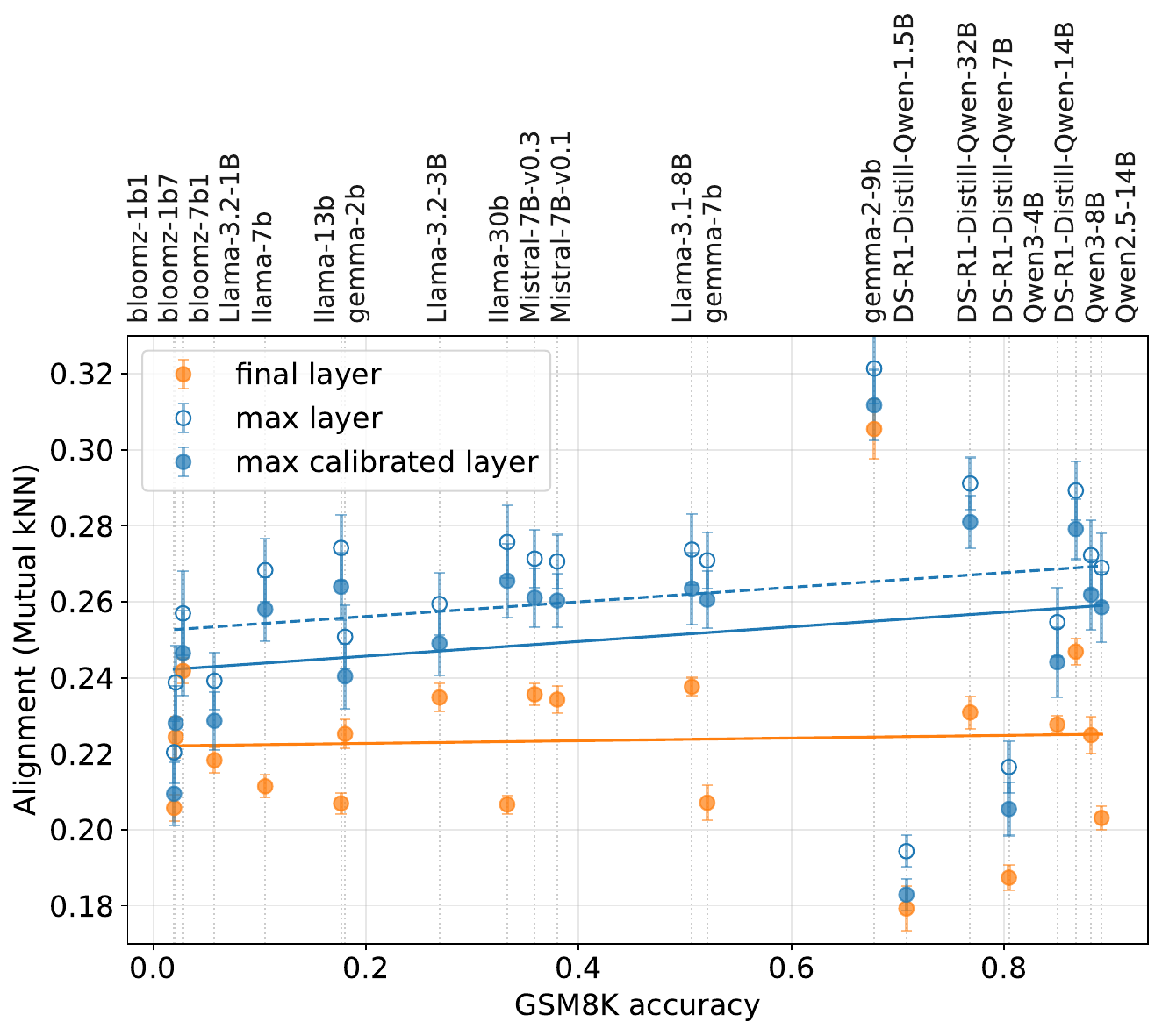}
\caption{GSM8K accuracy.}
\label{fig:gsm8k_alignment_docci_app}
\end{subfigure}
\caption{Performance of LLMs on ARC and GSM8K vs. alignment with vision models on DOCCI.}
\label{fig:arc_gsm8k_alignment_docci}
\end{figure*}

\subsection{Does higher alignment indicate stronger model performance?}
\label{subsec:performance-alignment}

The PRH originally suggested that stronger language models exhibit greater representational alignment with vision models on WIT \citep{huh2024platonic}. However, this trend does not extend to newer model families on the same dataset \citep{koepke2026back}. Following \citet{koepke2026back}, we examine the relationship between five-shot MMLU accuracy \citep{hendrycks2021measuringmassivemultitasklanguage} and alignment with DINOv3 across datasets with different caption characteristics. For each dataset, we independently sample 1024 image-caption pairs across five runs and report the mean and standard deviation of the mutual $k$-NN score with $k=10$. Since the original WIT subset provided by \citet{huh2024platonic} contains only 1024 samples, its evaluation corresponds to a single run.

Figure~\ref{fig:mmlu_alignment_four_datasets} compares the performance-alignment relationship on WIT, CC3M, and DOCCI using both original and summarized captions. We observe a weak to moderate positive correlation for maximally aligned layers on CC3M (Pearson $r=0.26$, $p=0.25$) and WIT ($r=0.46$, $p=0.03$), consistent with the findings of \citet{koepke2026back}. In contrast, DOCCI exhibits a stronger positive correlation when considering the maximally aligned layers ($r=0.66$, $p=0.001$), which remains significant with calibrated mutual $k$-NN ($r=0.66$, $p=0.001$). Importantly, this trend persists when DOCCI captions are compressed to 10\% of their original length, with $r=0.59$ ($p=0.005$) for the maximally aligned layers. Caption verbosity alone therefore does not explain the performance-alignment trend. Instead, the substantially weaker trend on the short and noisy WIT and CC3M captions, together with its persistence on original and summarized DOCCI captions, suggests that sufficiently clean and visually informative data is important for revealing shared structure between vision and language representations.

To assess whether these MMLU findings generalize to other measures of language-model capability, Figure~\ref{fig:arc_gsm8k_alignment_docci} relates alignment with DINOv3 on DOCCI to four-shot ARC Challenge accuracy \citep{clark2018arc} and five-shot GSM8K exact-match accuracy \citep{cobbe2021gsm8k}. ARC performance correlates strongly with alignment at the maximally aligned layers (Pearson $r=0.73$, $p=0.0002$), and the correlation remains significant after calibration (Pearson $r=0.73$, $p=0.0002$). This is in line with the MMLU results and shows that the performance-alignment trend is not specific to a single benchmark. Solving MMLU and ARC requires factual and semantic knowledge about the world that may be represented similarly by both vision and language models. In contrast, GSM8K primarily evaluates mathematical reasoning, which is less directly grounded in the semantic knowledge shared with visual representations, and its performance does not correlate significantly with alignment. Therefore, cross-modal alignment is not a universal proxy for language-model capability, but may be more closely associated with capabilities that rely on semantic knowledge. Results for ARC and GSM8K with alignment measured on the remaining caption datasets are provided in Appendix~\ref{apd:additional_performance-alignment_results}.

\subsection{How to interpret the magnitude of cross-modal alignment?}
\label{subsec:contextualization}

Alignment scores are difficult to interpret in isolation. A score such as 0.3 does not indicate by itself whether two representation spaces are weakly or strongly aligned. Therefore, we explicitly compare four regimes: randomly initialized models, pairs of unimodal vision and language models as the main object of study, explicitly aligned VLMs as cross-modal references, and pairs of unimodal models within one modality.

\begin{figure*}[t]
\centering
\centering
\includegraphics[width=1.0\columnwidth]{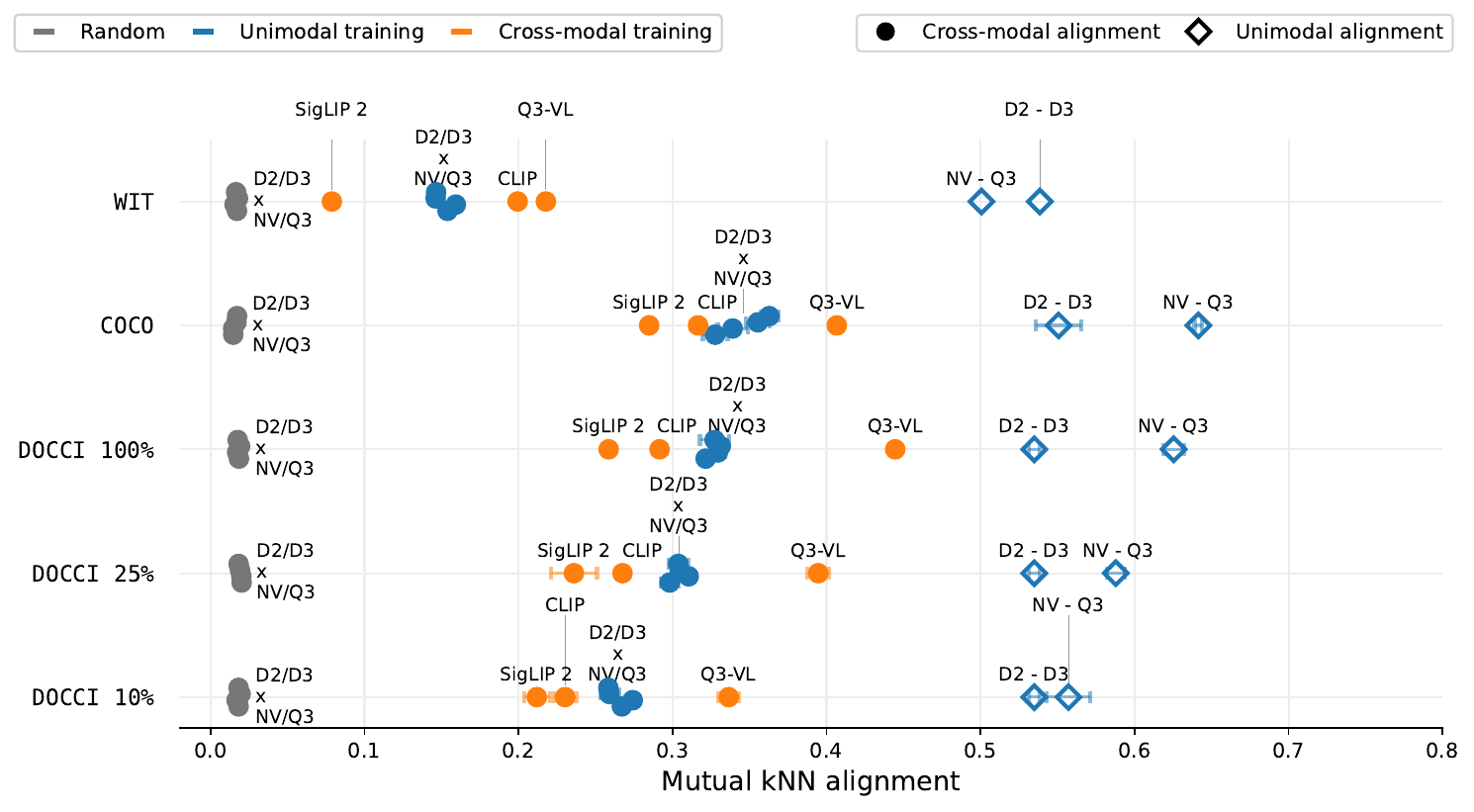}
\label{fig:alignment_scores}
\vspace{-0.4cm}
\caption{Alignment for different models and evaluation protocols: DINOv2/v3 (D2/D3), Qwen3-Emb-8B (Q3), NV-Embed-v2 (NV), CLIP, SigLIP 2, Qwen3-VL-Emb-8B (Q3-VL) (mean and standard deviation across 3 runs).}
\label{fig:alignment_scores_sorted}
\end{figure*}

As displayed in Figure~\ref{fig:alignment_scores_sorted}, the cross-modal alignment of unimodal models on COCO and DOCCI reaches a mutual $k$-NN score above 0.3, surpassing VLMs such as CLIP and SigLIP~2, but remaining below Qwen3-VL-Embedding-8B. For VLMs, we use the final embeddings in the shared image-text space, using the maximum across layer pairs instead yields nearly identical scores (Appendix~\ref{apd:layer_selection_vlm_app}). Notably, Qwen3-VL-Embedding-8B reports state-of-the-art performance on multimodal retrieval benchmarks and achieves the highest mutual $k$-NN cross-modal alignment. This shows that strong downstream retrieval performance is compatible with high cross-modal neighborhood agreement. The cross-modal alignment of unimodal models remains far below a mutual $k$-NN score of 0.5 observed for models of the same modality. This gap indicates that unimodal vision and language models learn partially shared representations that support cross-modal neighborhood agreement, but remain substantially less similar than models operating on the same modality.

The DOCCI caption variants provide a controlled view of how caption detail affects this comparison while keeping the underlying images fixed. Shortening the captions reduces absolute cross-modal alignment, but leaves the relative ordering of the model pairs largely unchanged. In particular, DINOv2/v3 paired with Qwen3-Embedding-8B or NV-Embed-v2 continue to surpass CLIP and SigLIP~2. Caption detail therefore affects the magnitude of alignment without changing our main contextual conclusions. Results across all DOCCI caption lengths are provided in Appendix~\ref{apd:contextualization_app}.

The high similarity between independently pretrained unimodal models is consistent with prior work: \citet{zhang2025sail} showed that their representations can be mapped into a shared embedding space using lightweight alignment layers and limited paired data, outperforming CLIP.

\section{Conclusion}

We revisited the PRH through a fine-grained analysis of alignment between unimodal vision and language models. Cross-modal alignment emerges in later model layers and is preserved under substantial dimensionality reduction. Moreover, we find that performance-alignment trends of the PRH persist for newer language models on image-text datasets with sufficiently informative captions.
Finally, our baseline analysis shows that unimodal vision and language models can reach the representational alignment of explicitly aligned VLMs, but remain far below models of the same modality.

\paragraph{Limitations.} The conclusions of our empirical work are limited to the evaluated models, datasets, and alignment metrics. We focus on cross-modal alignment between unimodal encoders from the vision and language domain; whether our findings generalize to other modalities remains open. Moreover, mutual $k$-NN captures only one notion of representational alignment, and its absolute scores depend on the metric hyperparameters, representation extraction choices, and evaluation data. We mitigate these dependencies by demonstrating that our findings persist with calibrated mutual $k$-NN and by interpreting the raw scores relative to a wide set of baselines.

\begin{ack}
This work was partially funded by the ERC (853489 - DEXIM) and the Alfried Krupp von Bohlen und Halbach Foundation, which we thank for their generous support. The authors gratefully acknowledge the scientific support and resources of the AI service infrastructure \textit{LRZ AI Systems} provided by the Leibniz Supercomputing Centre (LRZ) of the Bavarian Academy of Sciences and Humanities (BAdW), funded by Bayerisches Staatsministerium für Wissenschaft und Kunst (StMWK). The authors gratefully acknowledge the Gauss Centre for Supercomputing e.V. (www.gausscentre.eu) for funding this project by providing computing time on the Supercomputer JUPITER at Jülich Supercomputing Centre (JSC). This work was supported by the EuroHPC JU and the Gauss Centre for Supercomputing (GCS) through funding by the European Commission, German Federal Ministry of Research, Technology and Space (BMFTR), the Ministry of Culture and Science of the State of North Rhine-Westphalia (MKW). This work benefited from Hi! PARIS and State funding managed by the French National Research Agency (ANR) under the France 2030 program, reference ANR-23-IACL-0005.
\end{ack}

\bibliographystyle{plainnat}
\bibliography{egbib}

\appendix

\newpage
\section{Broader impacts}
\label{apd:broader_impacts}
This work complements prior research on the Platonic Representation Hypothesis by identifying where alignment emerges across layers and dimensions, showing that its correlation with model performance re-emerges for sufficiently clean and informative captions, and contextualizing its magnitude against a broad set of baselines. While our study is diagnostic and does not introduce any new model, our findings may help practitioners select encoders and representations for efficient cross-modal alignment, potentially reducing paired supervision and computation. The evaluated models and image-text datasets may encode societal and cultural biases that aggregate alignment scores cannot reveal. Therefore, downstream applications will require task-specific evaluation.

\section{Layer-wise alignment analysis}
\label{apd:layer-wise_analysis}

\begin{figure*}[h]
\centering

\begin{subfigure}{0.32\textwidth}
    \centering
    \includegraphics[width=\linewidth]{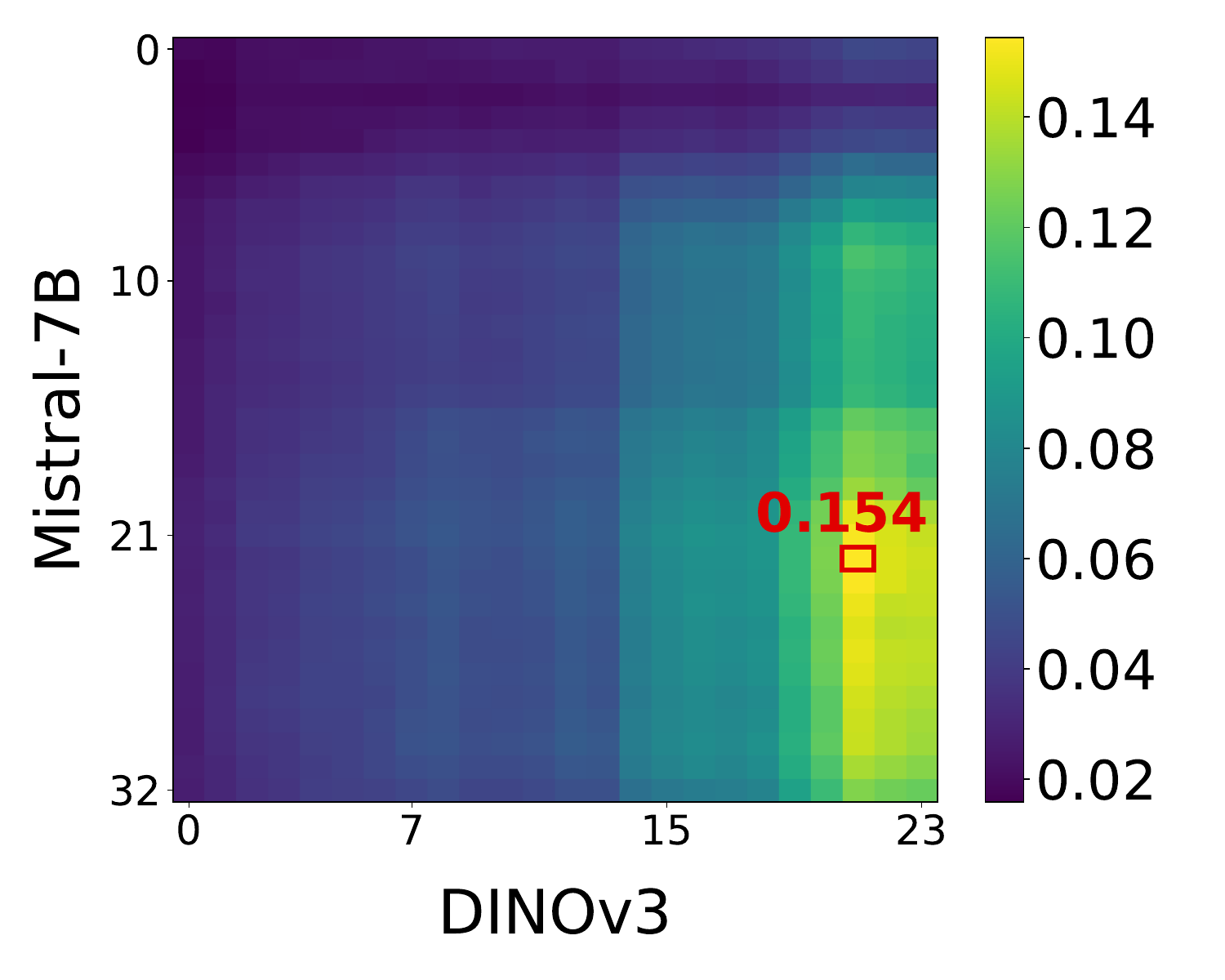}
\end{subfigure}
\hfill
\begin{subfigure}{0.32\textwidth}
    \centering
    \includegraphics[width=\linewidth]{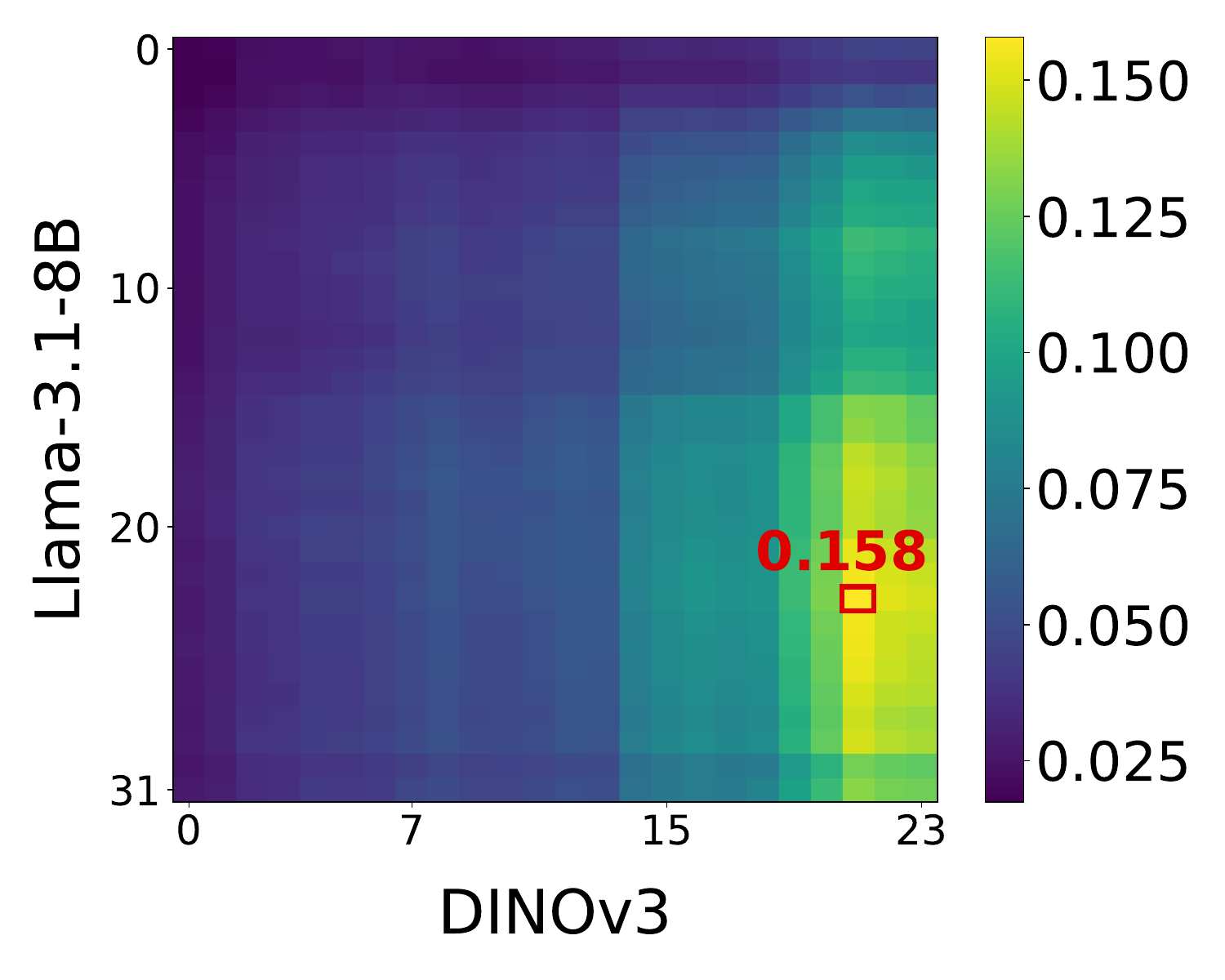}
\end{subfigure}
\hfill
\begin{subfigure}{0.32\textwidth}
    \centering
    \includegraphics[width=\linewidth]{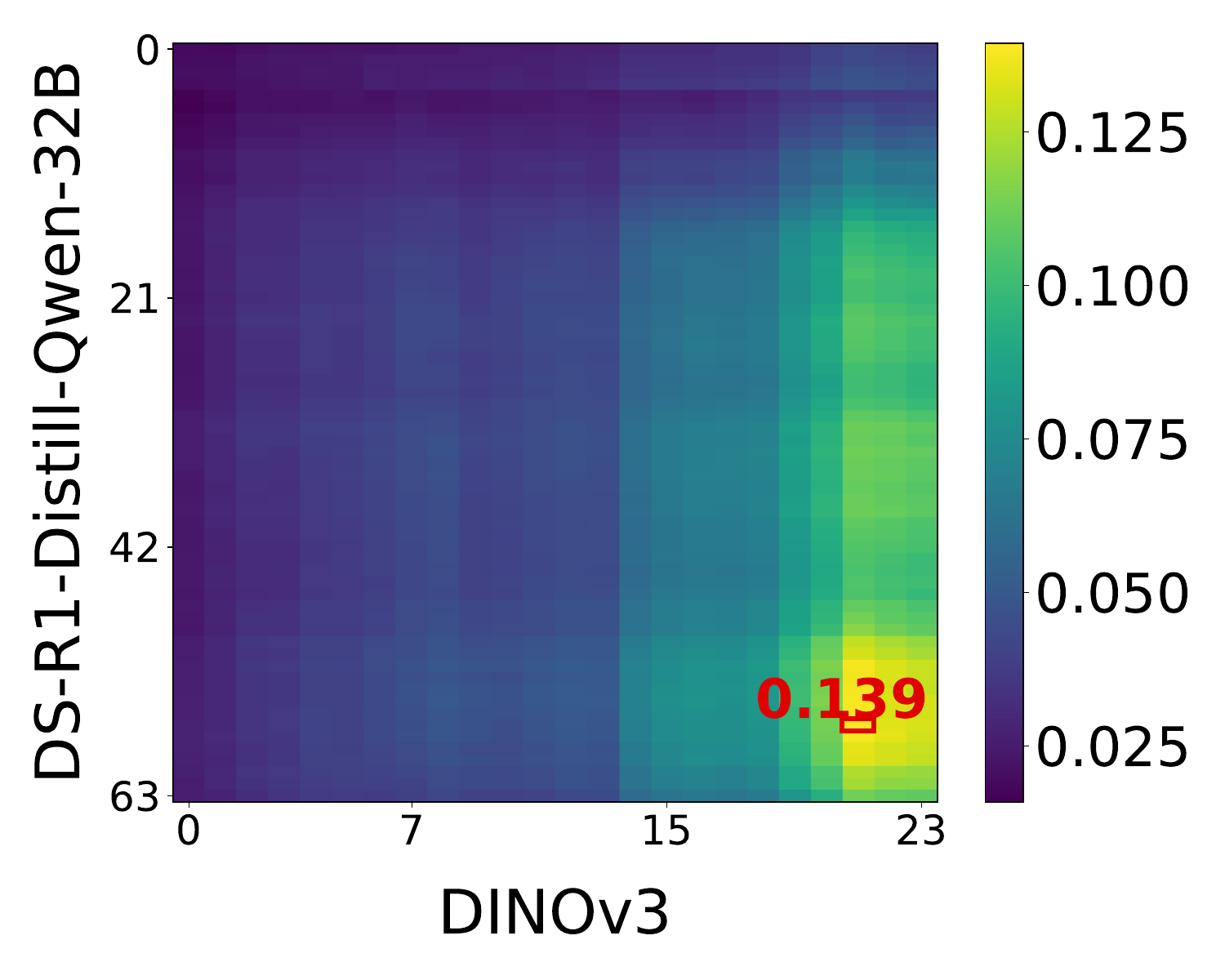}
\end{subfigure}

\caption*{Vision-Language}

\vspace{1em}

\begin{subfigure}{0.32\textwidth}
    \centering
    \includegraphics[width=\linewidth]{figures/heatmaps/vision-vision/DINOv2_DINOv3_wit_1024.pdf}
\end{subfigure}
\hfill
\begin{subfigure}{0.32\textwidth}
    \centering
    \includegraphics[width=\linewidth]{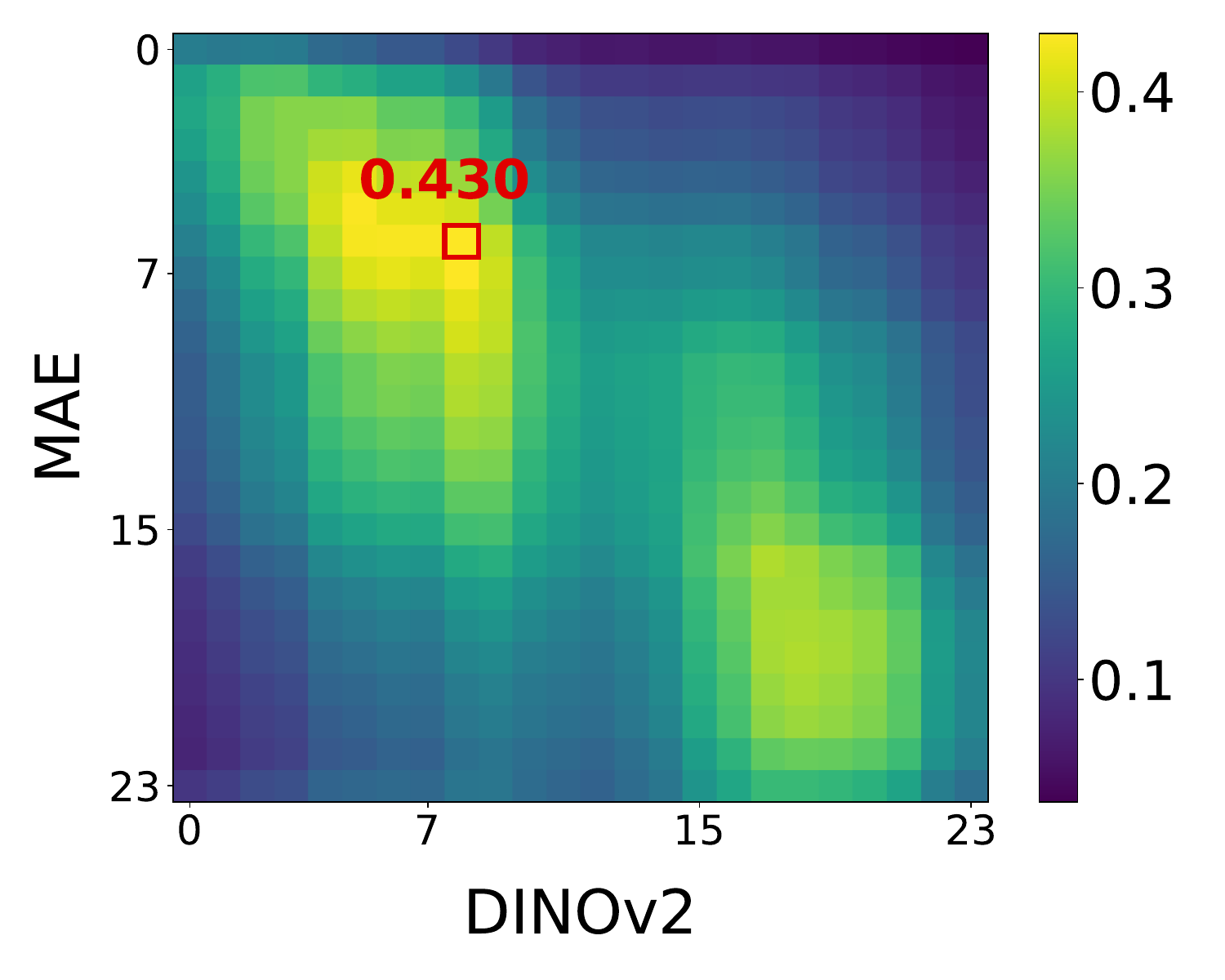}
\end{subfigure}
\hfill
\begin{subfigure}{0.32\textwidth}
    \centering
    \includegraphics[width=\linewidth]{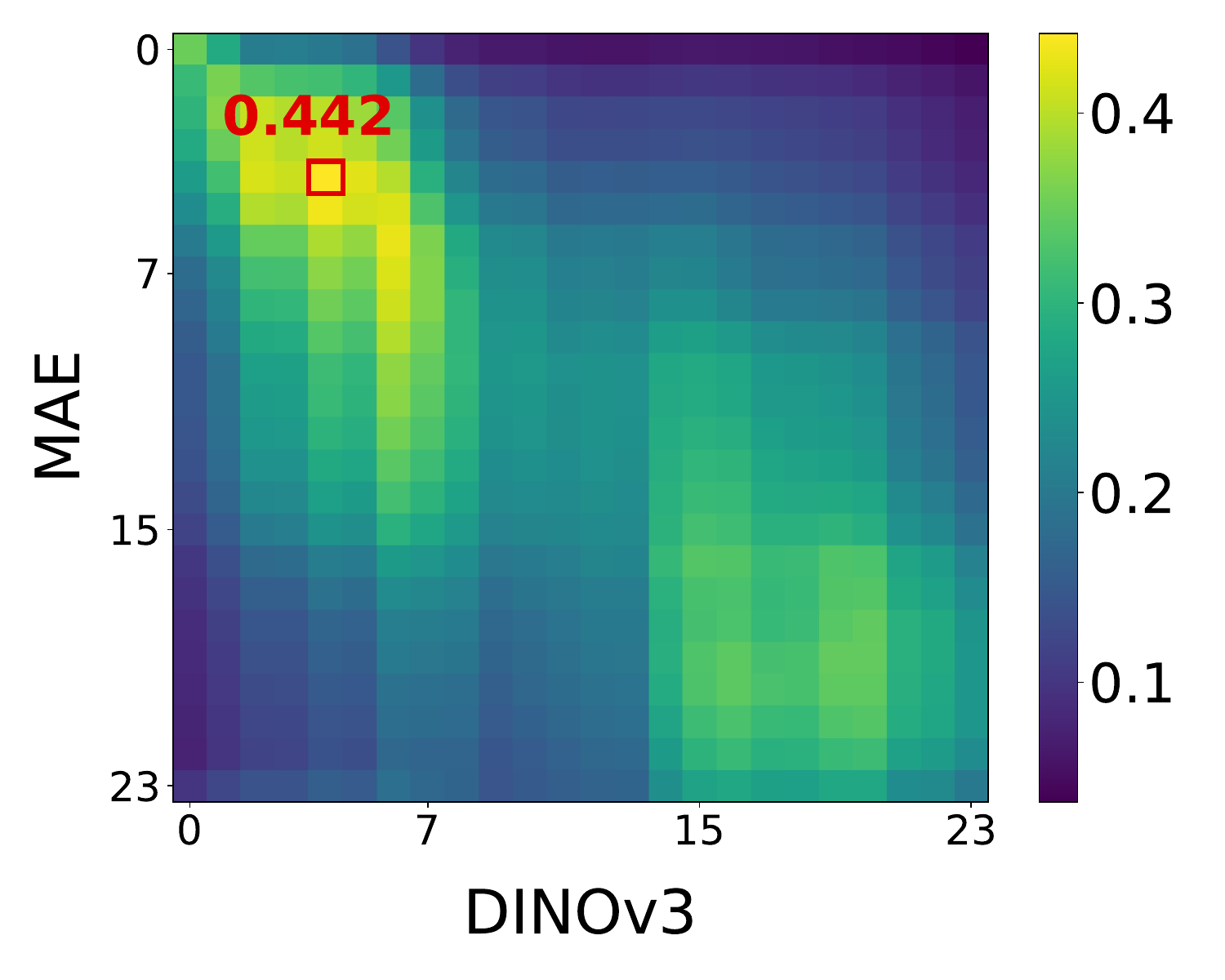}
\end{subfigure}

\caption*{Vision-Vision}

\vspace{1em}

\begin{subfigure}{0.32\textwidth}
    \centering
    \includegraphics[width=\linewidth]{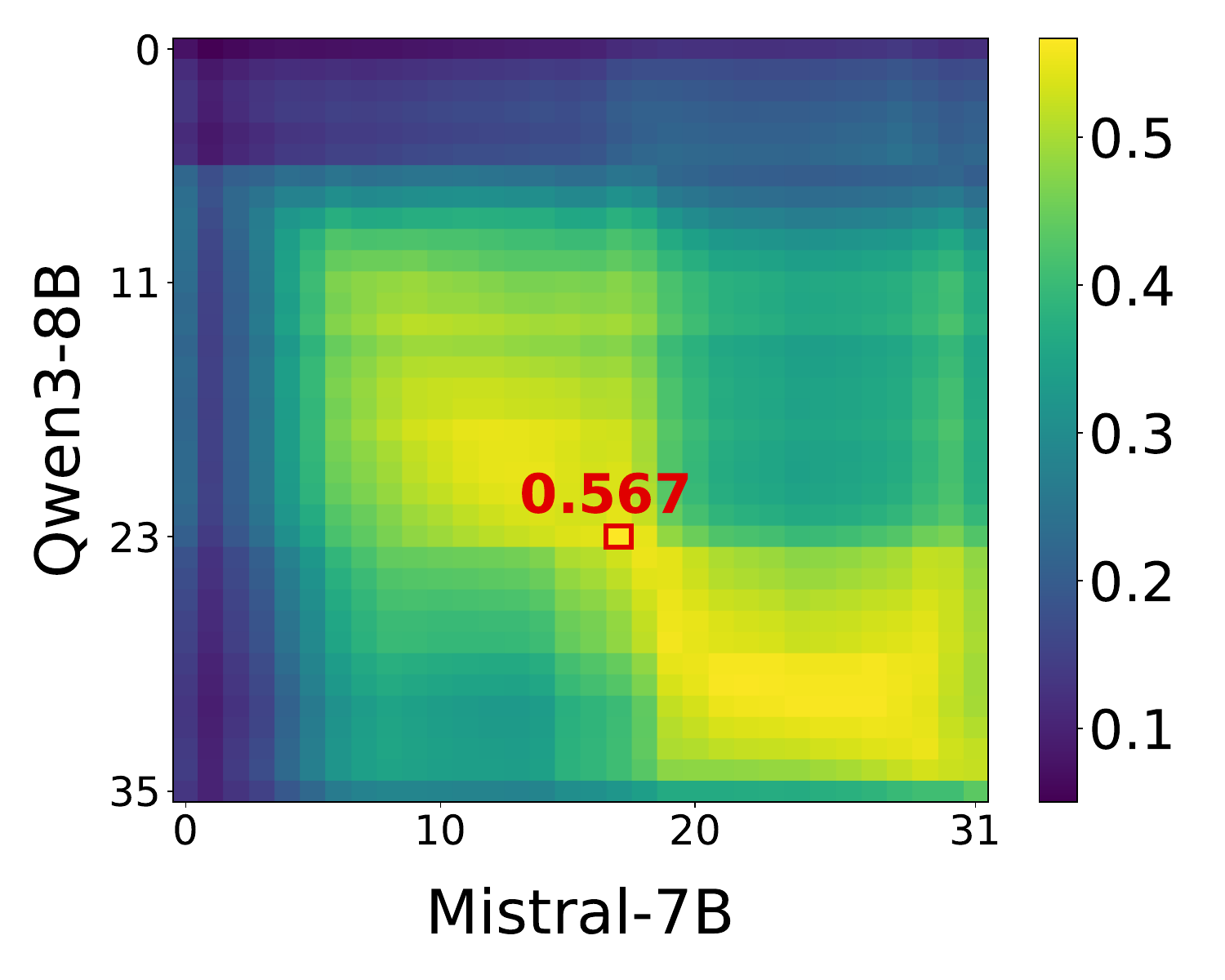}
\end{subfigure}
\hfill
\begin{subfigure}{0.32\textwidth}
    \centering
    \includegraphics[width=\linewidth]{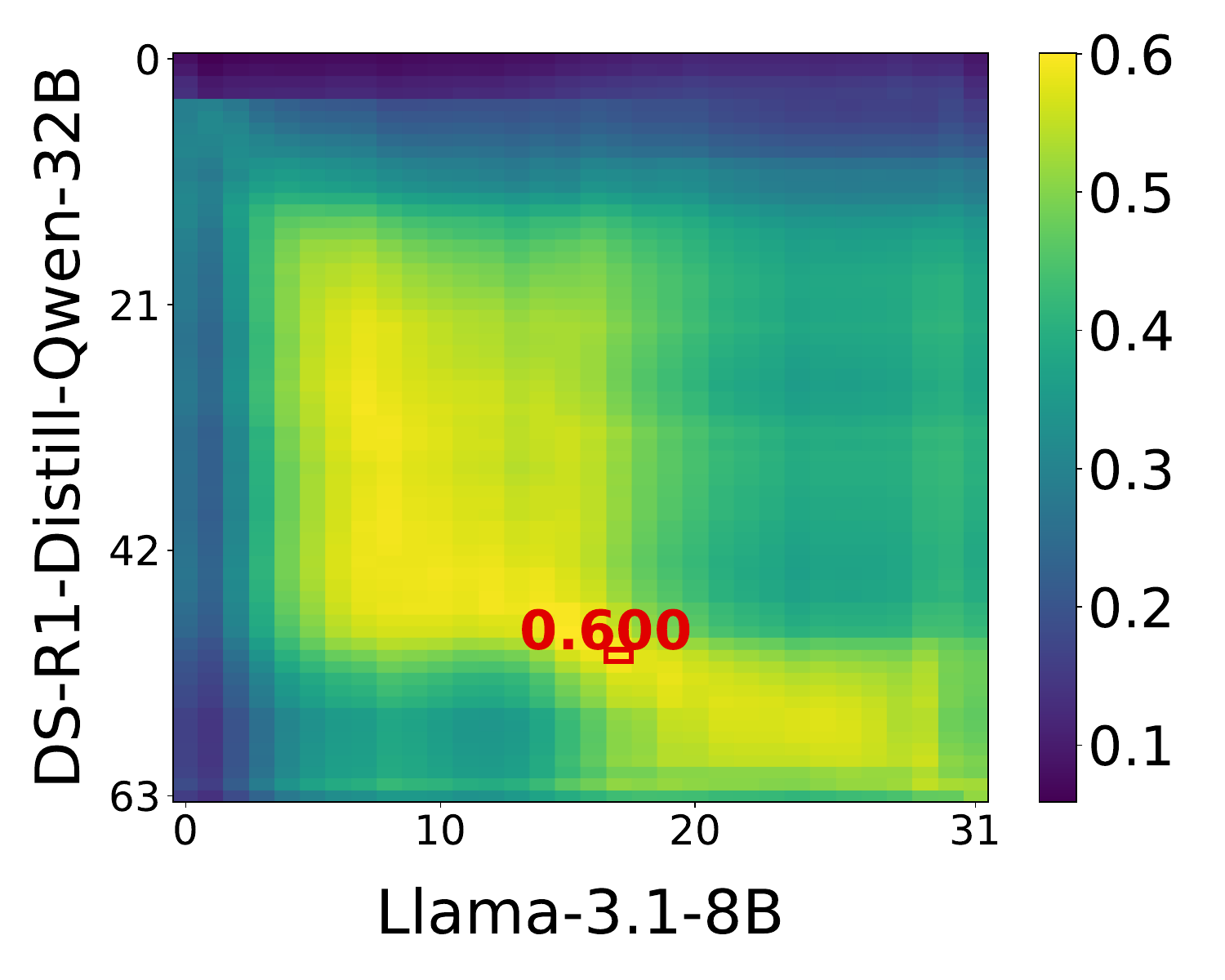}
\end{subfigure}
\hfill
\begin{subfigure}{0.32\textwidth}
    \centering
    \includegraphics[width=\linewidth]{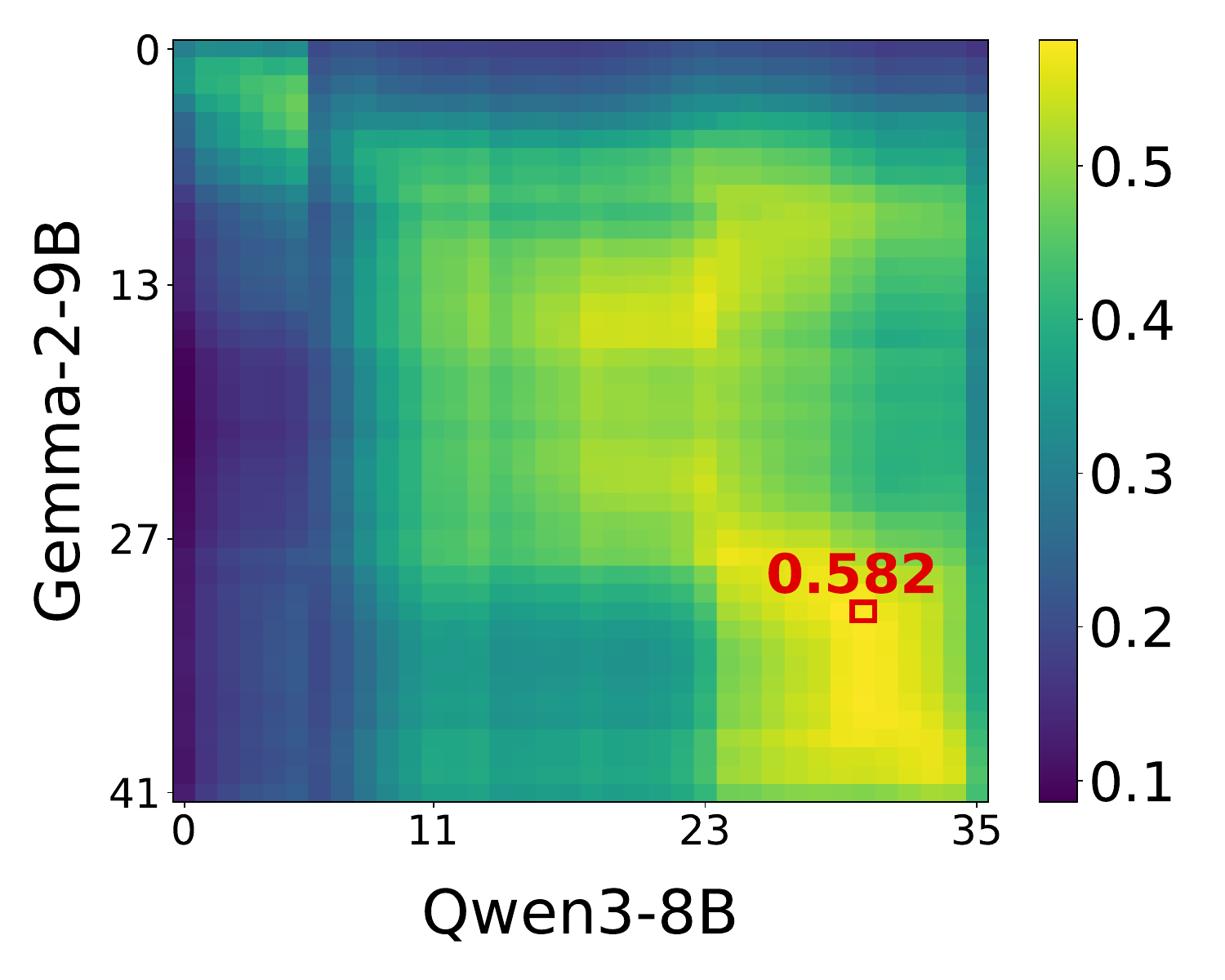}
\end{subfigure}

\caption*{Language-Language}

\caption{Heatmaps with mutual $k$-NN alignment between every layer on WIT.}
\label{fig:heatmap_alignment_layers_additional}
\end{figure*}

\begin{figure*}[h]
\centering

\begin{subfigure}{0.32\textwidth}
    \centering
    \includegraphics[width=\linewidth]{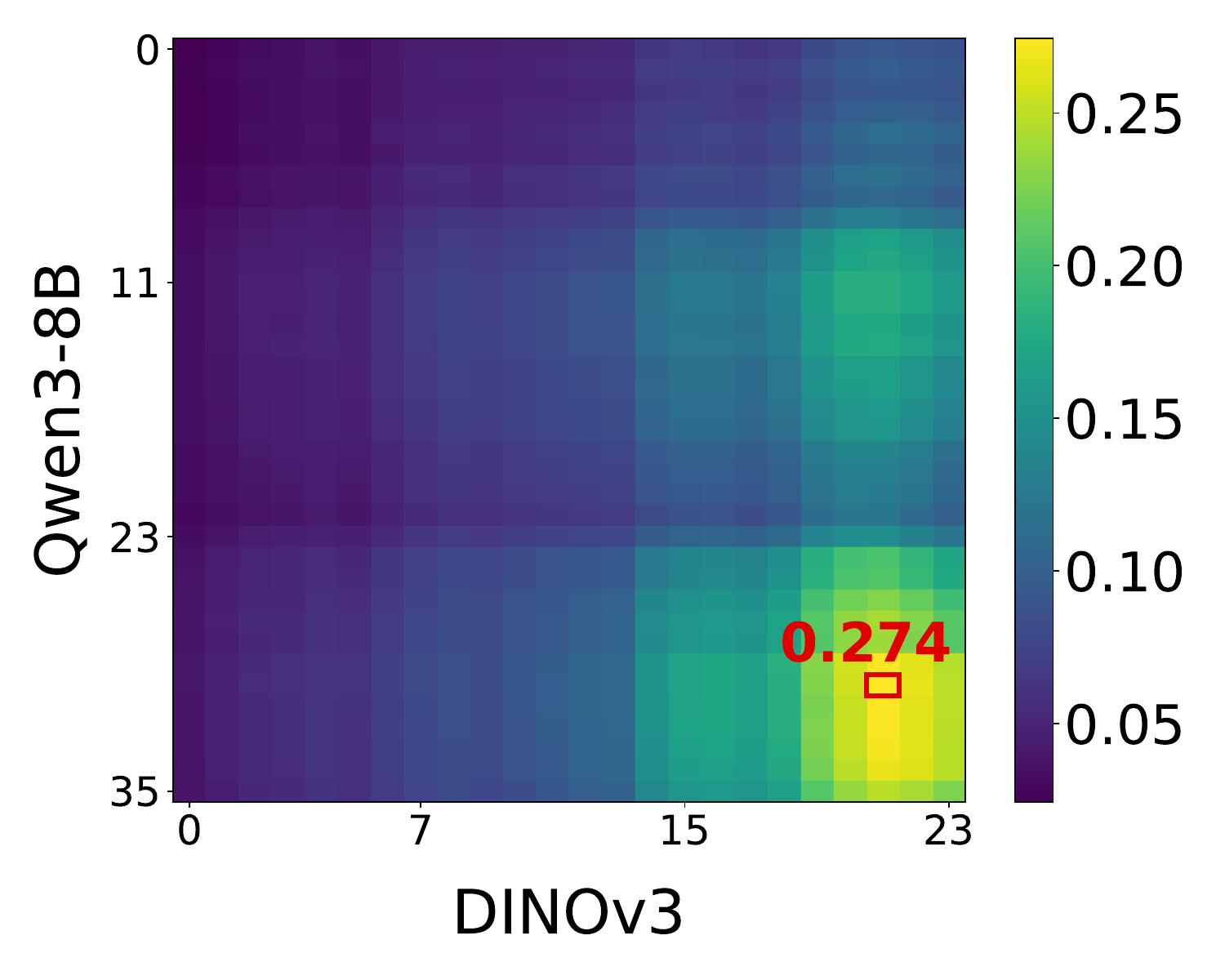}
\end{subfigure}
\hfill
\begin{subfigure}{0.32\textwidth}
    \centering
    \includegraphics[width=\linewidth]{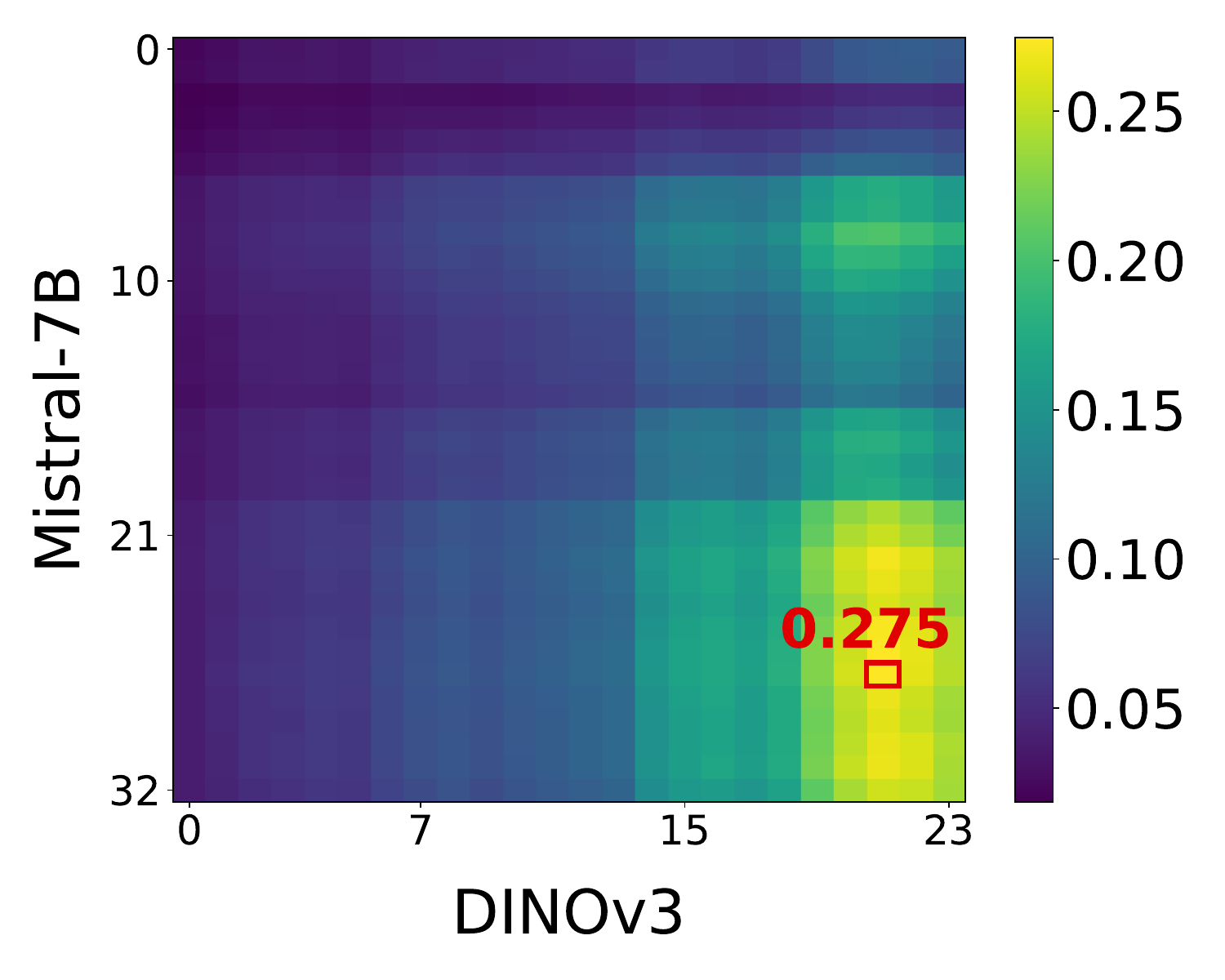}
\end{subfigure}
\hfill
\begin{subfigure}{0.32\textwidth}
    \centering
    \includegraphics[width=\linewidth]{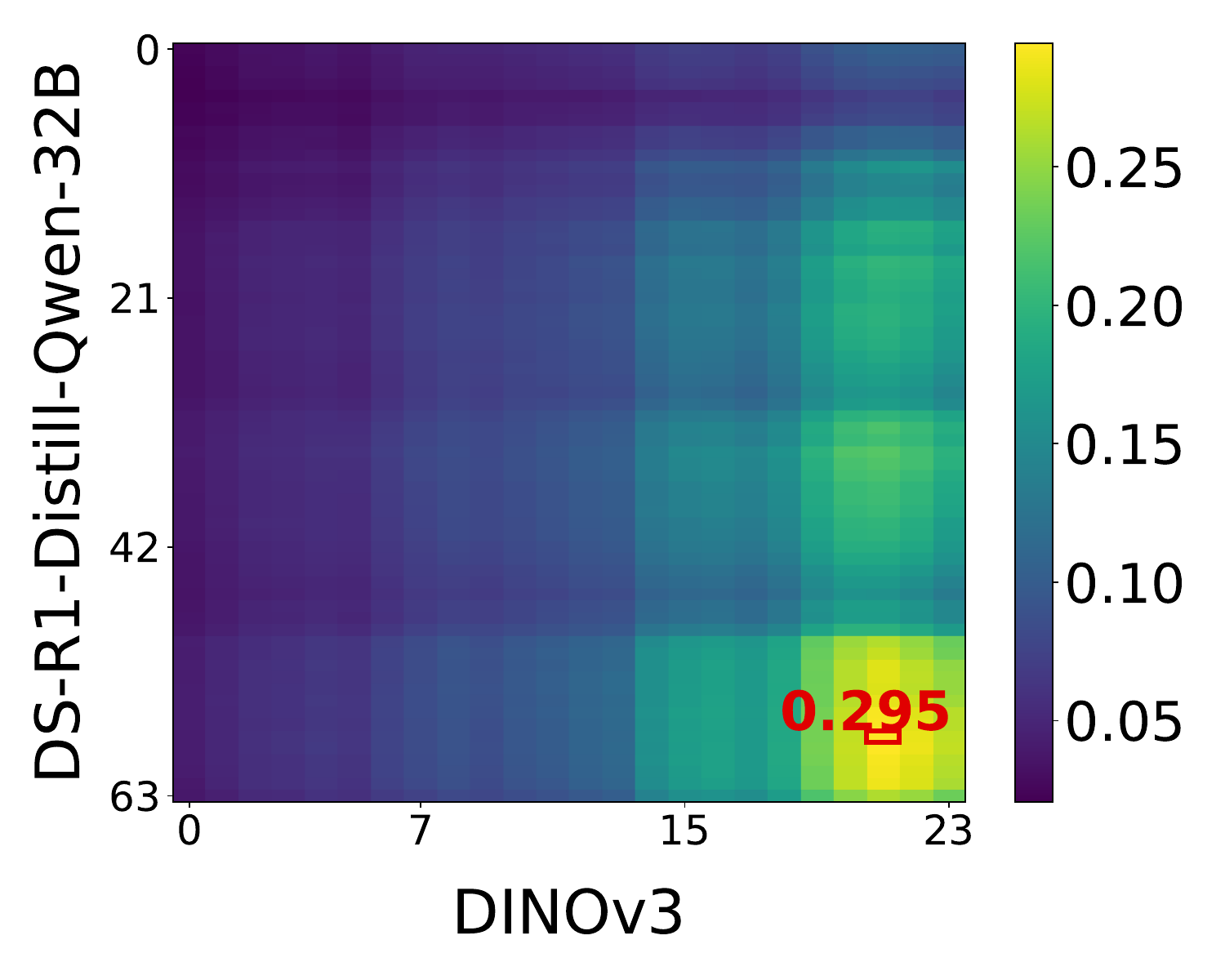}
\end{subfigure}

\caption*{Vision-Language}

\vspace{1em}

\begin{subfigure}{0.32\textwidth}
    \centering
    \includegraphics[width=\linewidth]{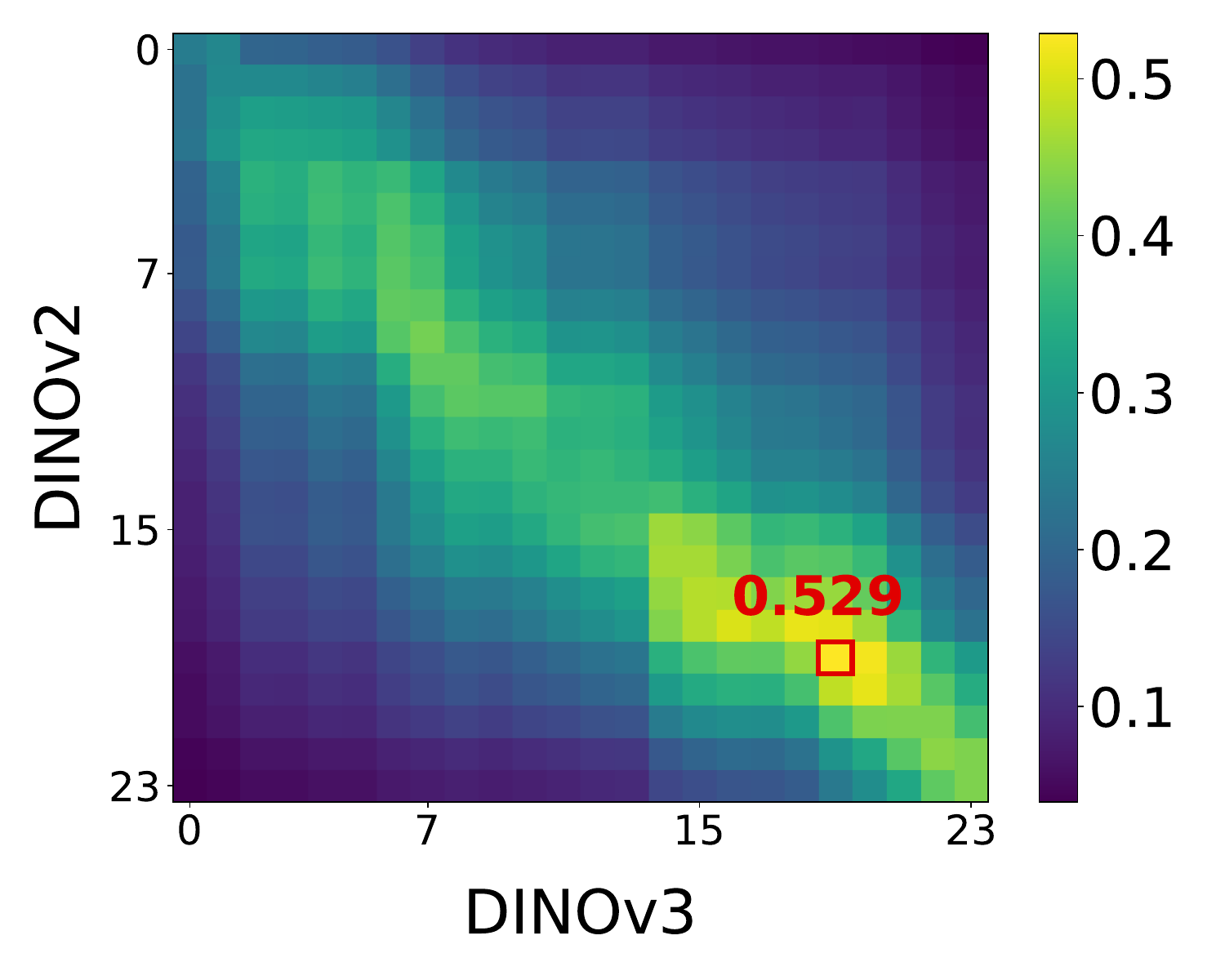}
\end{subfigure}
\hfill
\begin{subfigure}{0.32\textwidth}
    \centering
    \includegraphics[width=\linewidth]{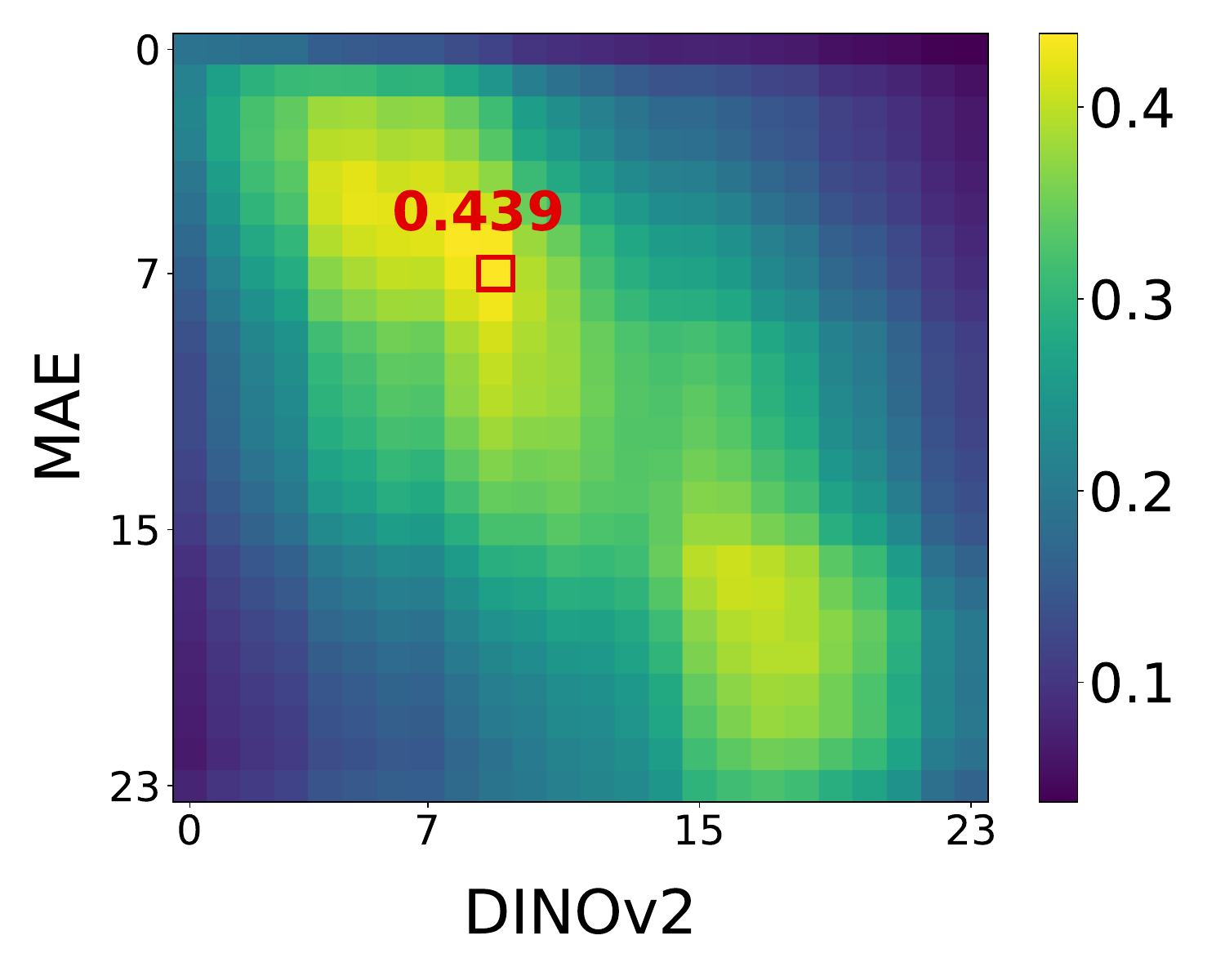}
\end{subfigure}
\hfill
\begin{subfigure}{0.32\textwidth}
    \centering
    \includegraphics[width=\linewidth]{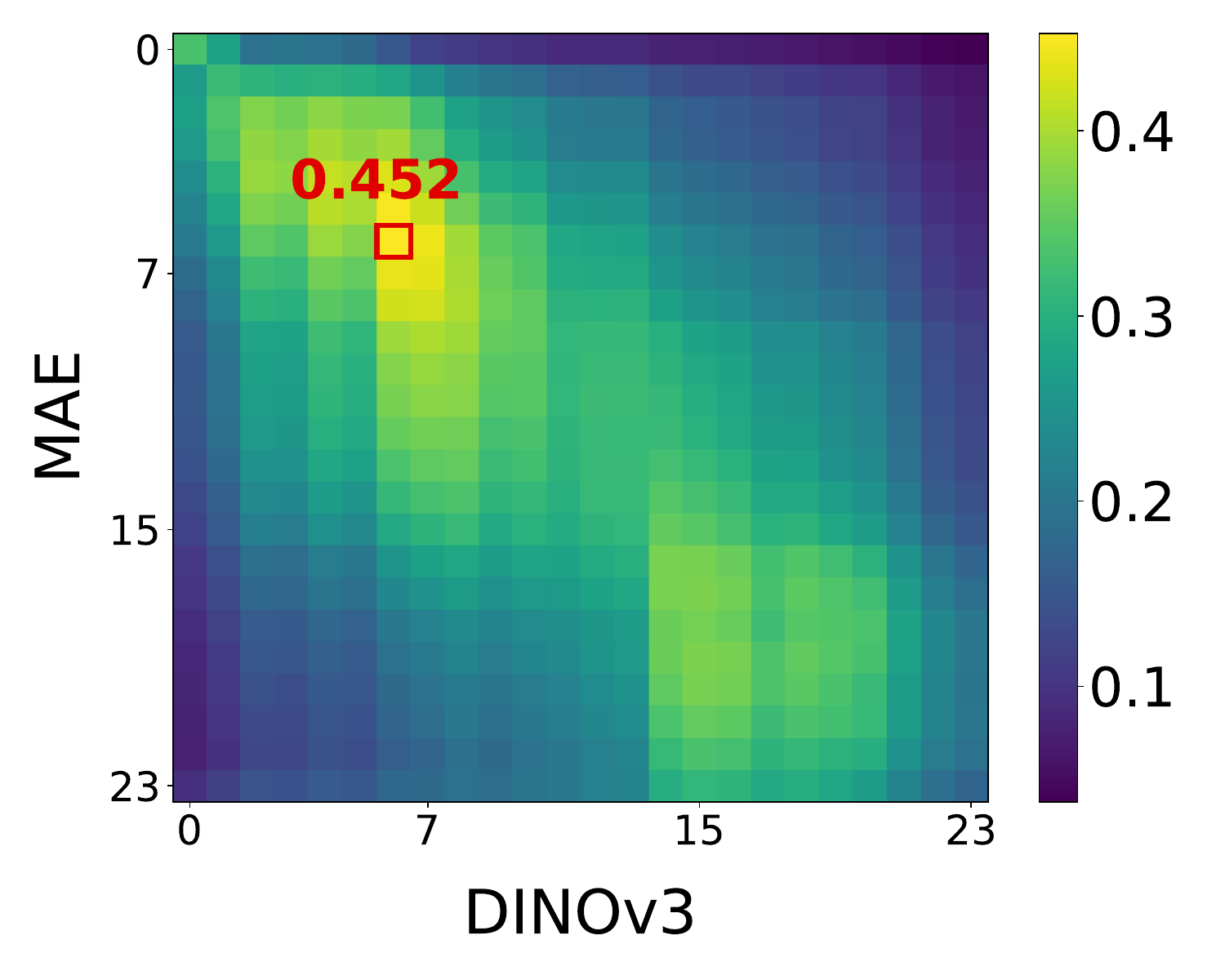}
\end{subfigure}

\caption*{Vision-Vision}

\vspace{1em}

\begin{subfigure}{0.32\textwidth}
    \centering
    \includegraphics[width=\linewidth]{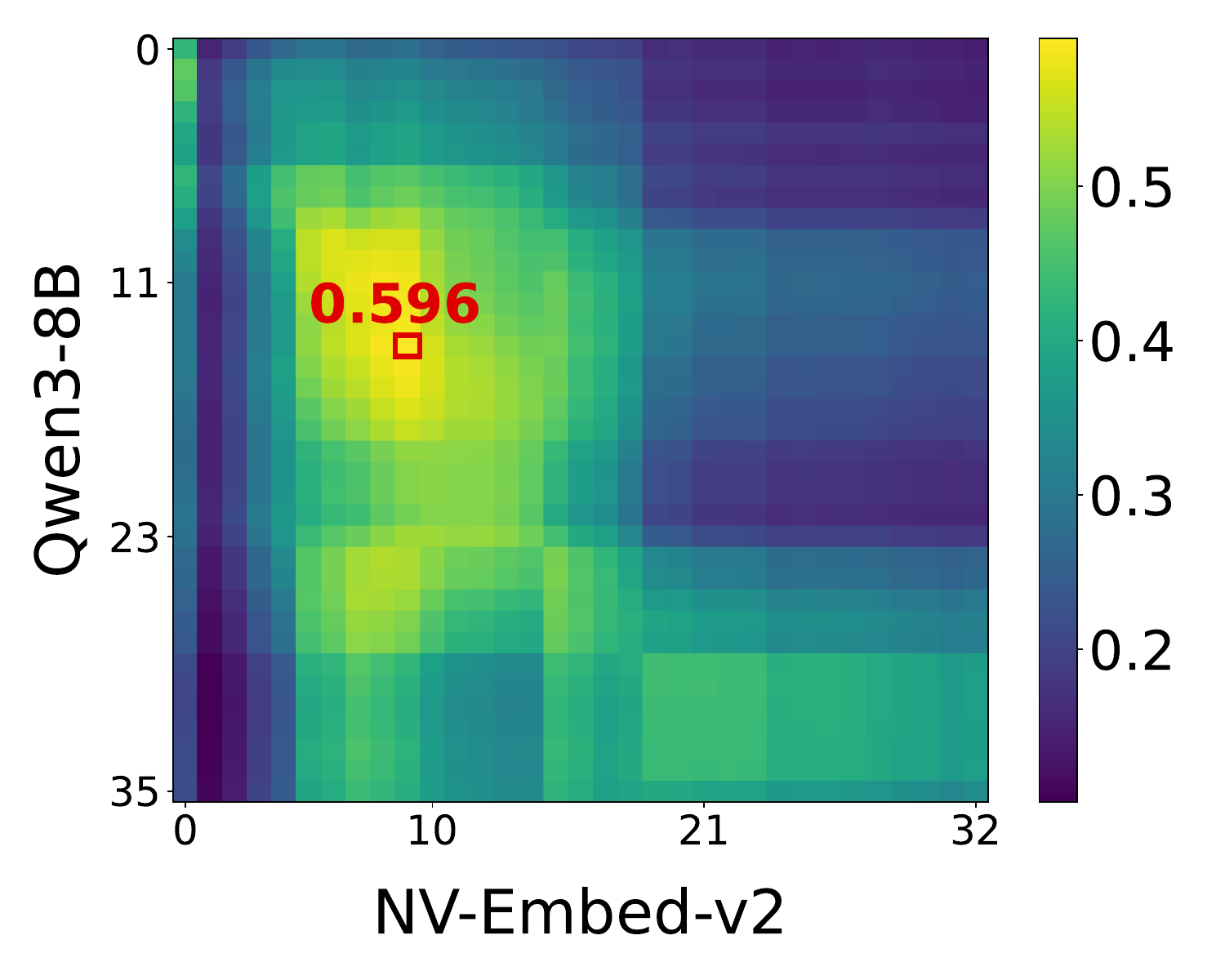}
\end{subfigure}
\hfill
\begin{subfigure}{0.32\textwidth}
    \centering
    \includegraphics[width=\linewidth]{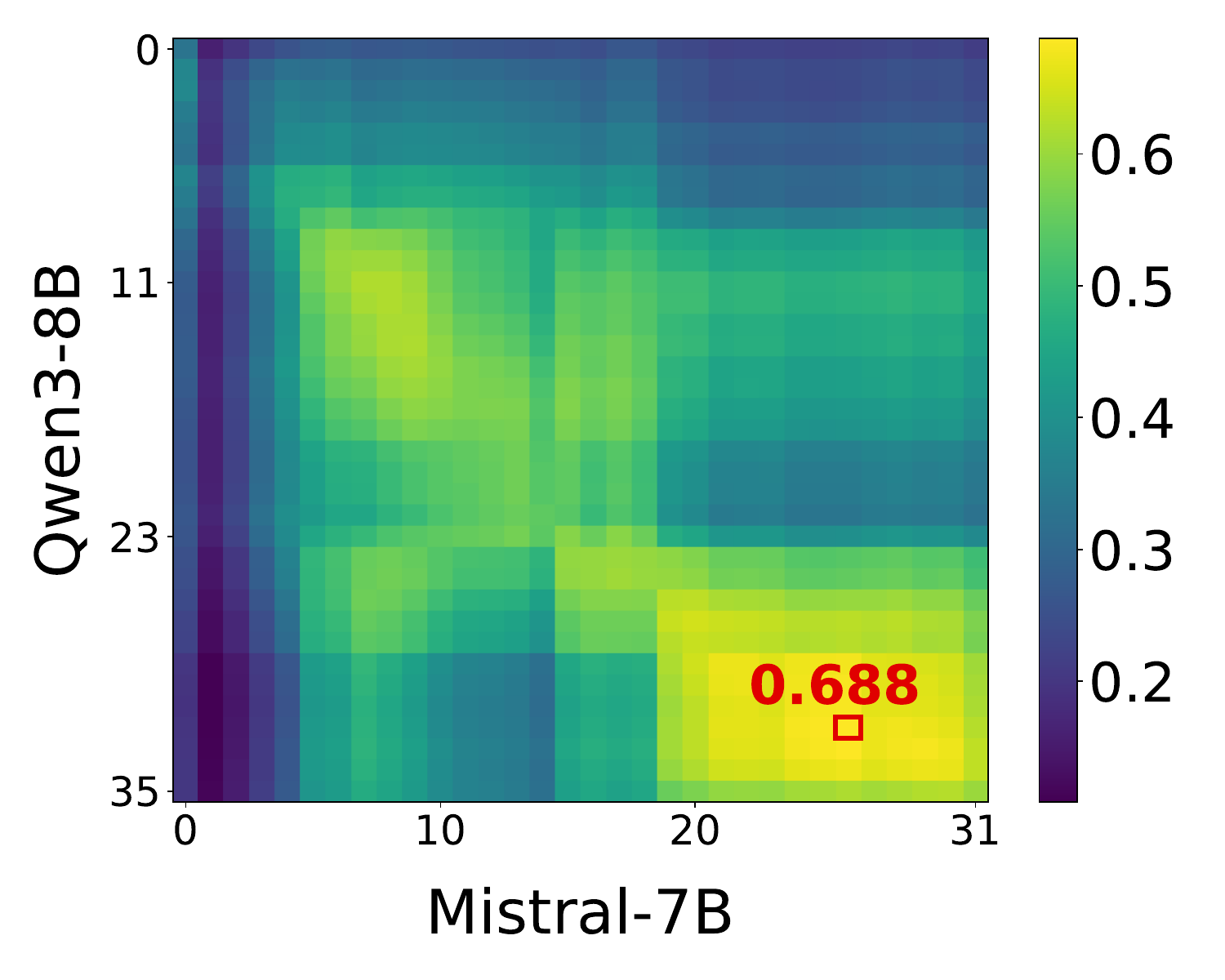}
\end{subfigure}
\hfill
\begin{subfigure}{0.32\textwidth}
    \centering
    \includegraphics[width=\linewidth]{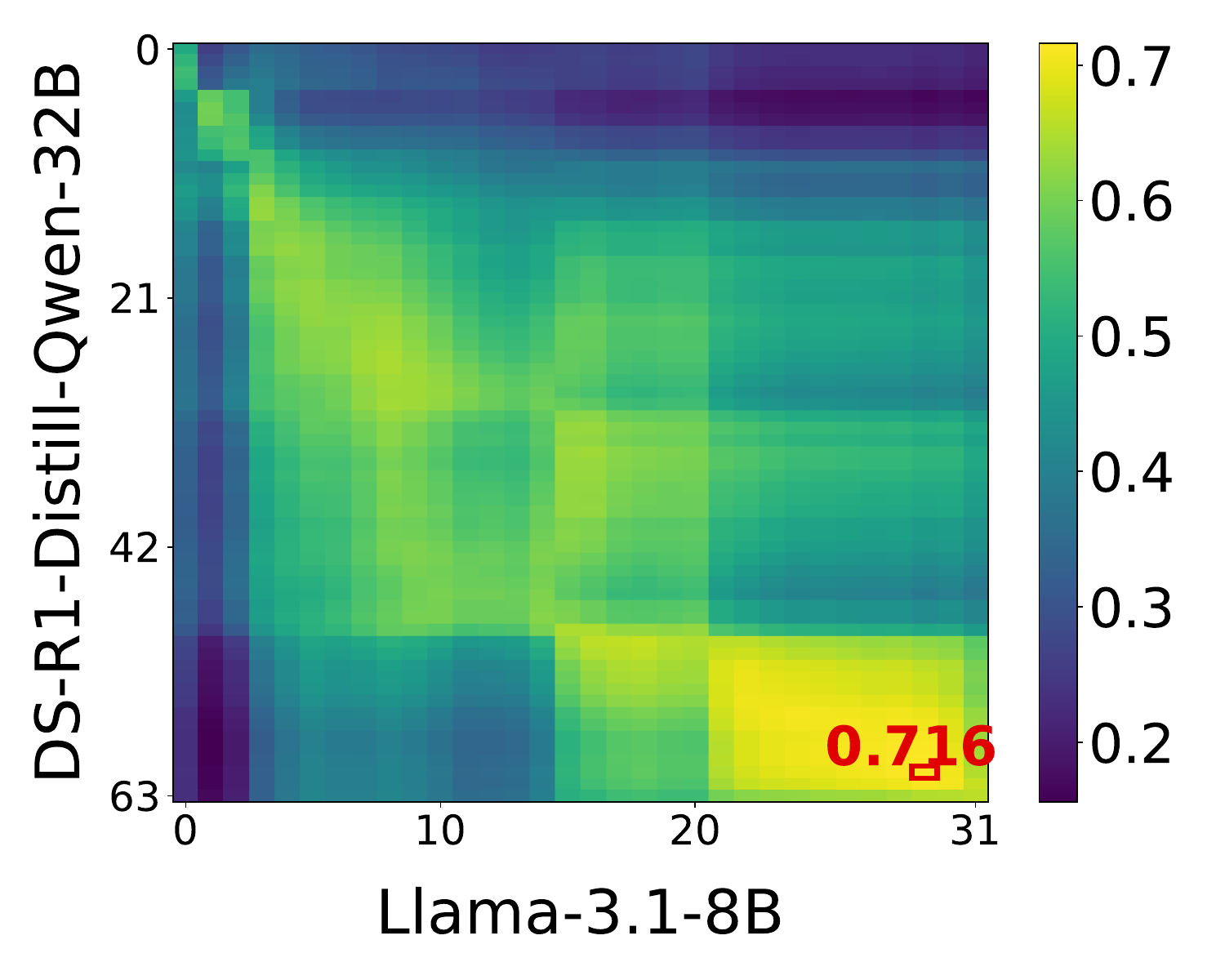}
\end{subfigure}

\caption*{Language-Language}

\caption{Heatmaps with mutual $k$-NN alignment between every layer on DOCCI.}
\label{fig:heatmap_alignment_layers_additional_docci}
\end{figure*}

Complementing the analysis in Section~\ref{subsec:where}, we provide additional heatmaps with layer-wise alignment scores for more model combinations on WIT in Figure~\ref{fig:heatmap_alignment_layers_additional} and on DOCCI in Figure~\ref{fig:heatmap_alignment_layers_additional_docci}. Here, we also include layer-wise alignment scores between two generative language models for the language-language regime. While mutual k-NN alignment between a generative model and a language embedding model is primarily concentrated in early-to-intermediate layer pairs, alignment between two generative language models often peaks in later layers.

\clearpage
\section{Compression of representations via PCA}
\label{apd:compression_pca}
\begin{figure}[ht]
    \centering
    \begin{subfigure}[t]{0.48\linewidth}
        \centering
        \includegraphics[width=\linewidth]{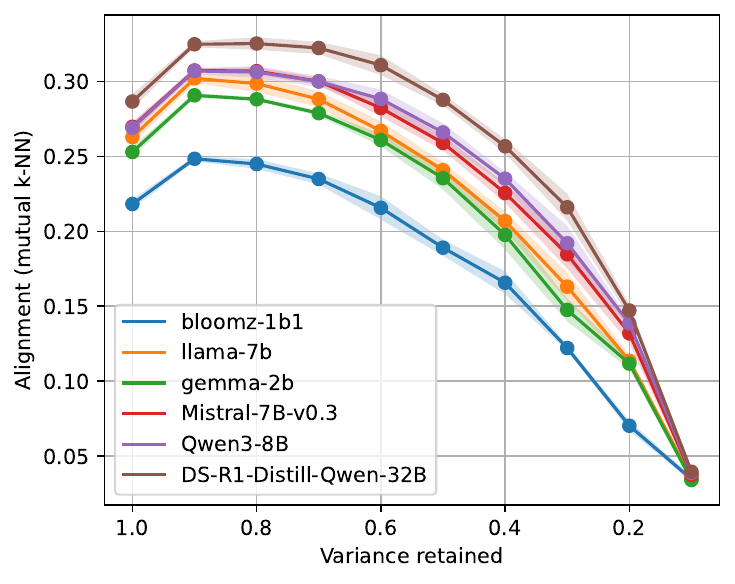}
        \caption{Explained variance.}
        \label{fig:compression_var_docci}
    \end{subfigure}
    \hfill
    \begin{subfigure}[t]{0.48\linewidth}
        \centering
        \includegraphics[width=\linewidth]{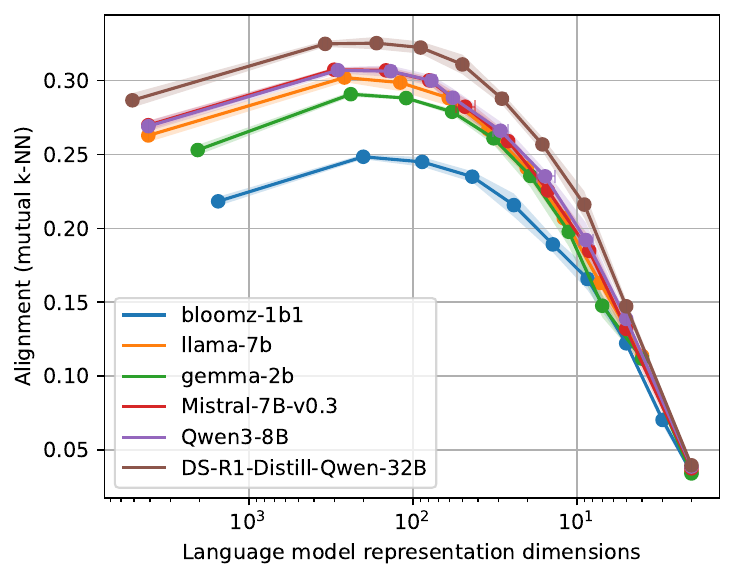}
        \caption{Number of dimensions.}
        \label{fig:compression_dims_docci}
    \end{subfigure}
    \caption{Effect of PCA compression on alignment between DINOv2 and language models on DOCCI.}
    \label{fig:compression_docci}
\end{figure}

\begin{figure}[ht]
    \centering
    \begin{subfigure}[t]{0.48\linewidth}
        \centering
        \includegraphics[width=\linewidth]{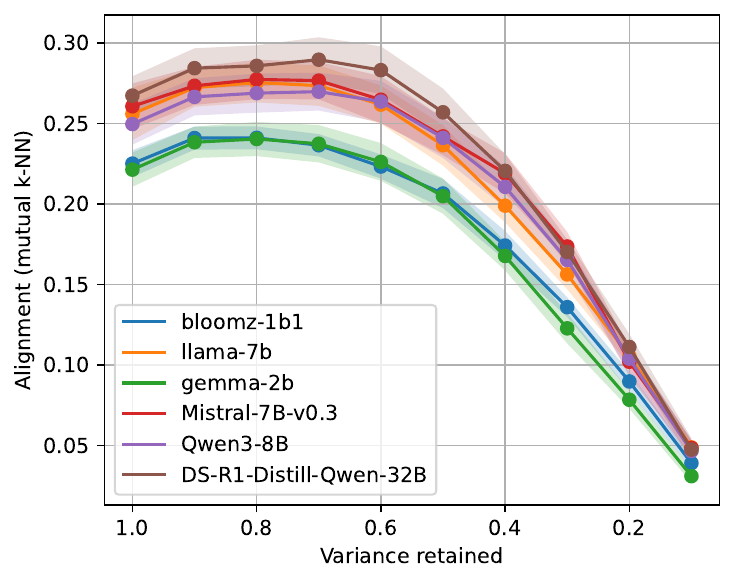}
        \caption{Explained variance.}
        \label{fig:compression_var_coco}
    \end{subfigure}
    \hfill
    \begin{subfigure}[t]{0.48\linewidth}
        \centering
        \includegraphics[width=\linewidth]{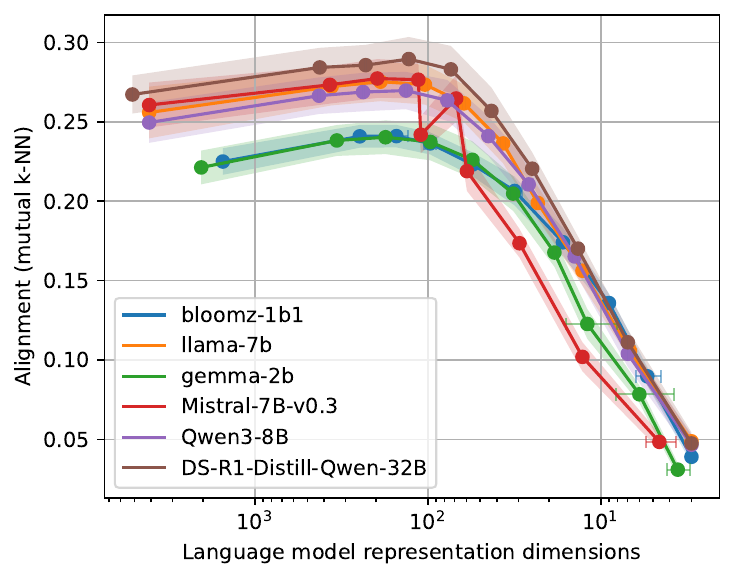}
        \caption{Number of dimensions.}
        \label{fig:compression_dims_coco}
    \end{subfigure}
    \caption{Effect of PCA compression on alignment between DINOv2 and language models on COCO.}
    \label{fig:compression_coco}
\end{figure}

Figure~\ref{fig:compression_docci} extends the PCA compression analysis from Section~\ref{subsec:where} to DOCCI. We report the mean and standard deviation across 5 runs. Figure~\ref{fig:compression_var_docci} shows that, on DOCCI, a larger fraction of the variance must be retained to preserve alignment, while alignment remains largely preserved when compressing the language representations to 100 dimensions (Figure~\ref{fig:compression_dims_docci}). 
Figure~\ref{fig:compression_coco} shows the results on COCO.

\clearpage
\section{Performance-alignment trends}
\label{apd:additional_performance-alignment_results}

\begin{figure*}[ht]
\centering
\begin{subfigure}{0.48\linewidth}
\centering
\includegraphics[width=\linewidth]{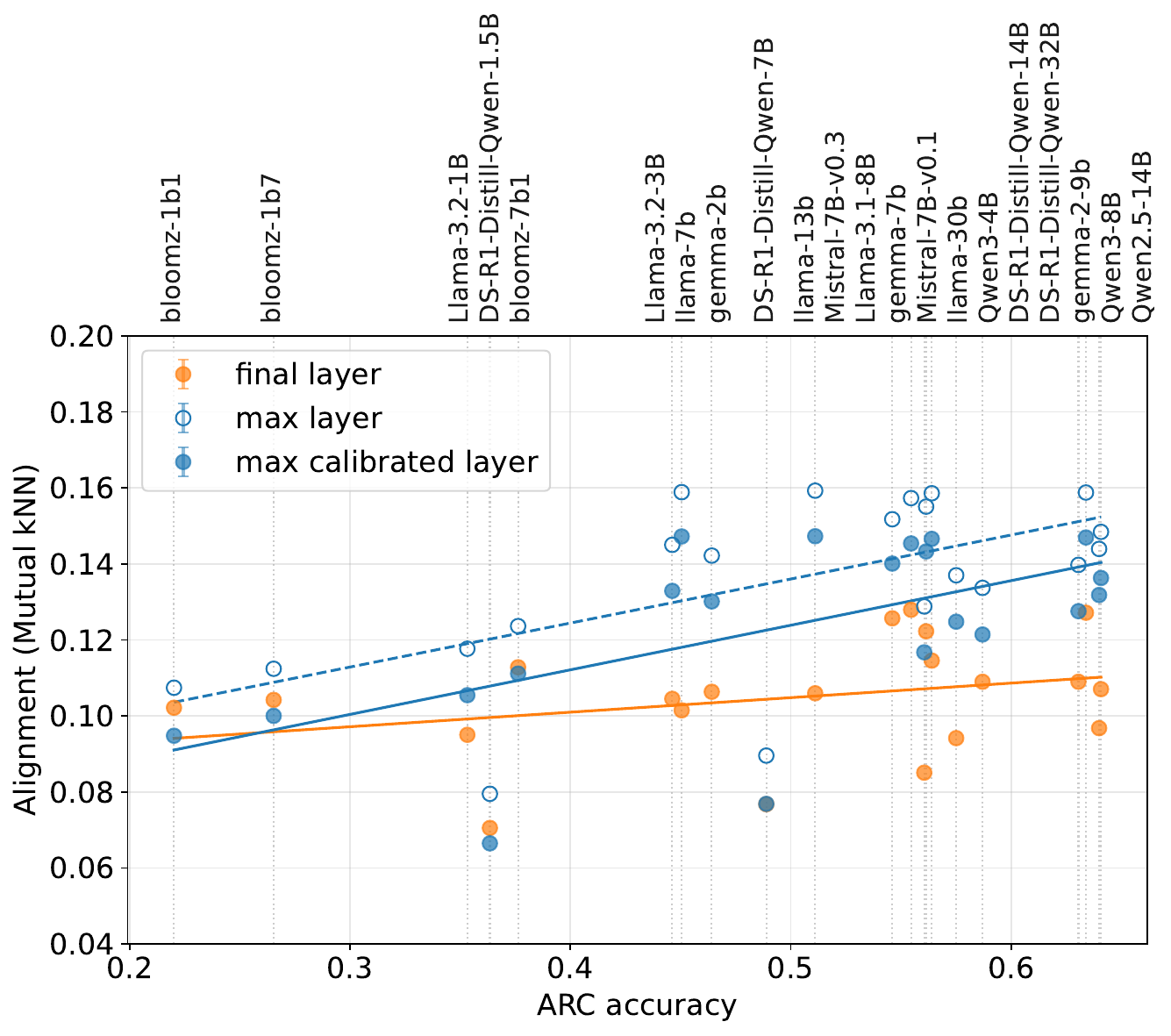}
\caption{WIT.}
\label{fig:arc_alignment_wit_app}
\end{subfigure}
\hfill
\begin{subfigure}{0.48\linewidth}
\centering
\includegraphics[width=\linewidth]{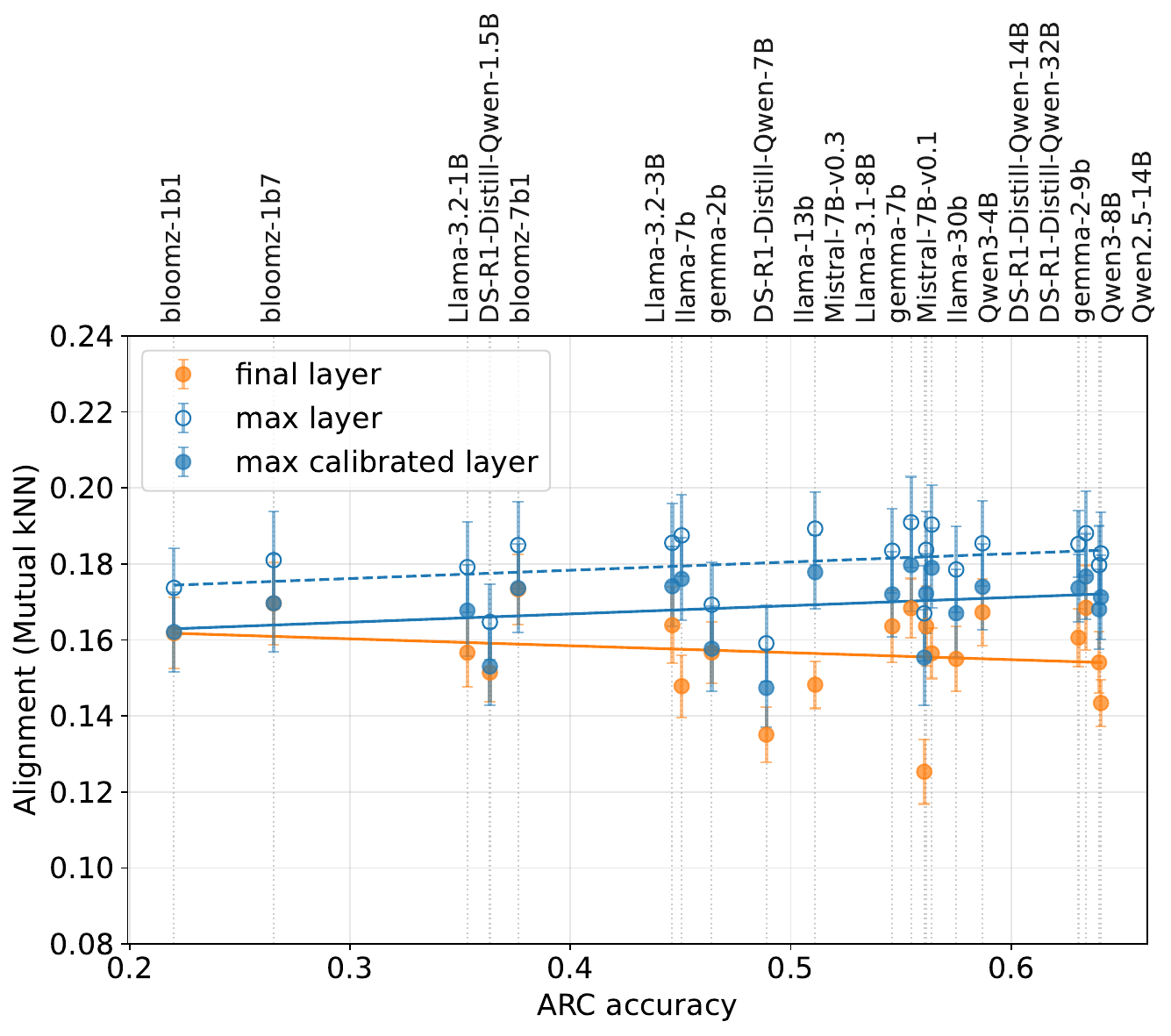}
\caption{CC3M short synthetic.}
\label{fig:arc_alignment_cc3m_short_app}
\end{subfigure}
\begin{subfigure}{0.48\linewidth}
\centering
\includegraphics[width=\linewidth]{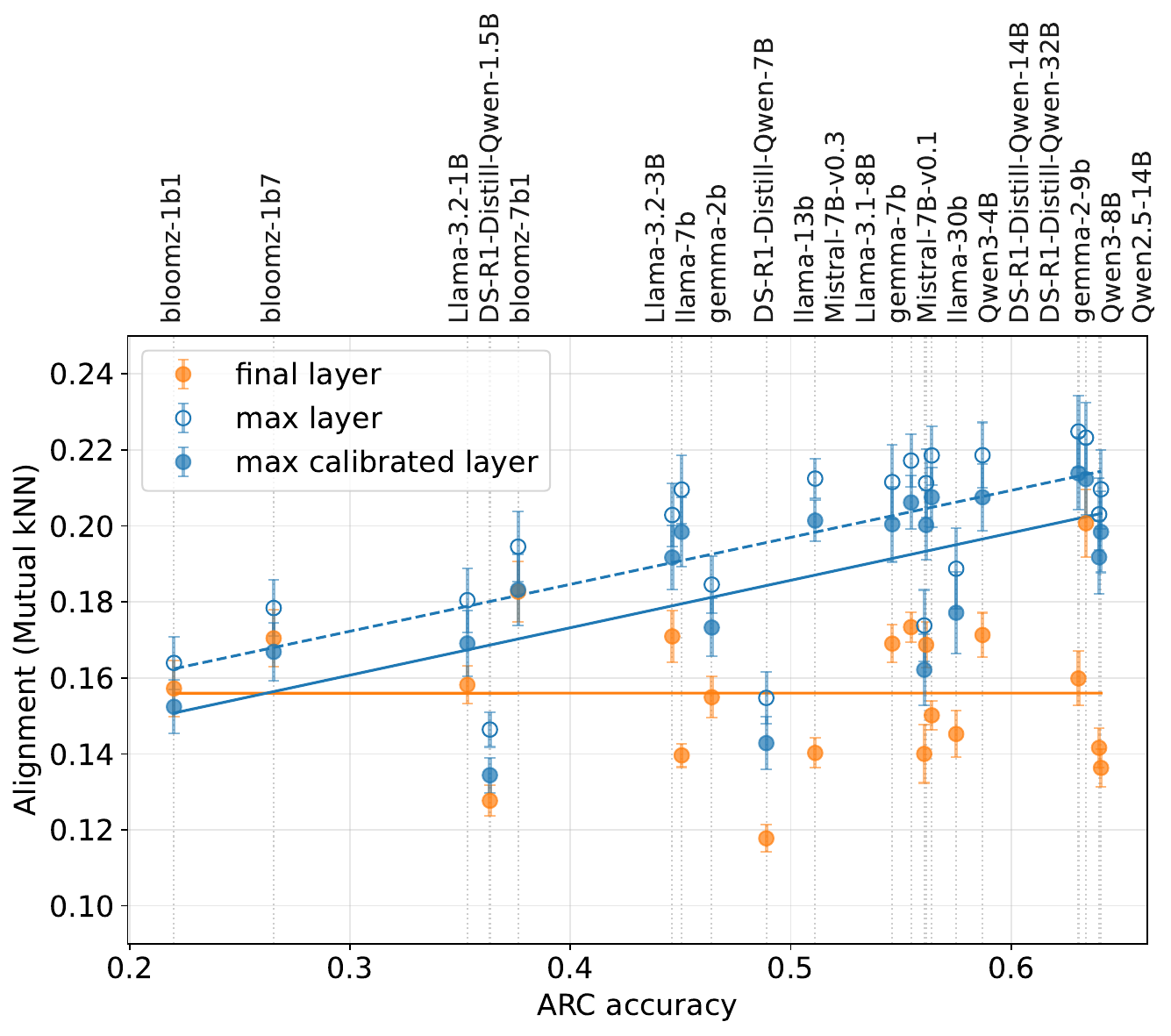}
\caption{DOCCI short synthetic.}
\label{fig:arc_alignment_docci_short_app}
\end{subfigure}
\hfill
\begin{subfigure}{0.48\linewidth}
\centering
\includegraphics[width=\linewidth]{figures/mutual_knn_arc_with_calibration_final/dinov3/arc_challenge_mutual_knn_docci_with_summaries_description.pdf}
\caption{DOCCI.}
\label{fig:arc_alignment_docci_long_app}
\end{subfigure}
\caption{Performance on ARC vs.\ alignment with DINOv3.}
\label{fig:arc_alignment_four_datasets}
\end{figure*}

Section~\ref{subsec:performance-alignment} analyzes the relationship between language-model performance on ARC and GSM8K and cross-modal alignment only on DOCCI. Figure~\ref{fig:arc_alignment_four_datasets} extends the ARC analysis to WIT, CC3M, and summarized DOCCI captions. The positive relationship observed on original DOCCI remains visible, although more weakly, on WIT and summarized DOCCI when considering the maximally aligned layers. In contrast, CC3M exhibits little consistent relationship, and final-layer alignment is generally less predictive.

\begin{figure*}[ht]
\centering
\begin{subfigure}{0.48\linewidth}
\centering
\includegraphics[width=\linewidth]{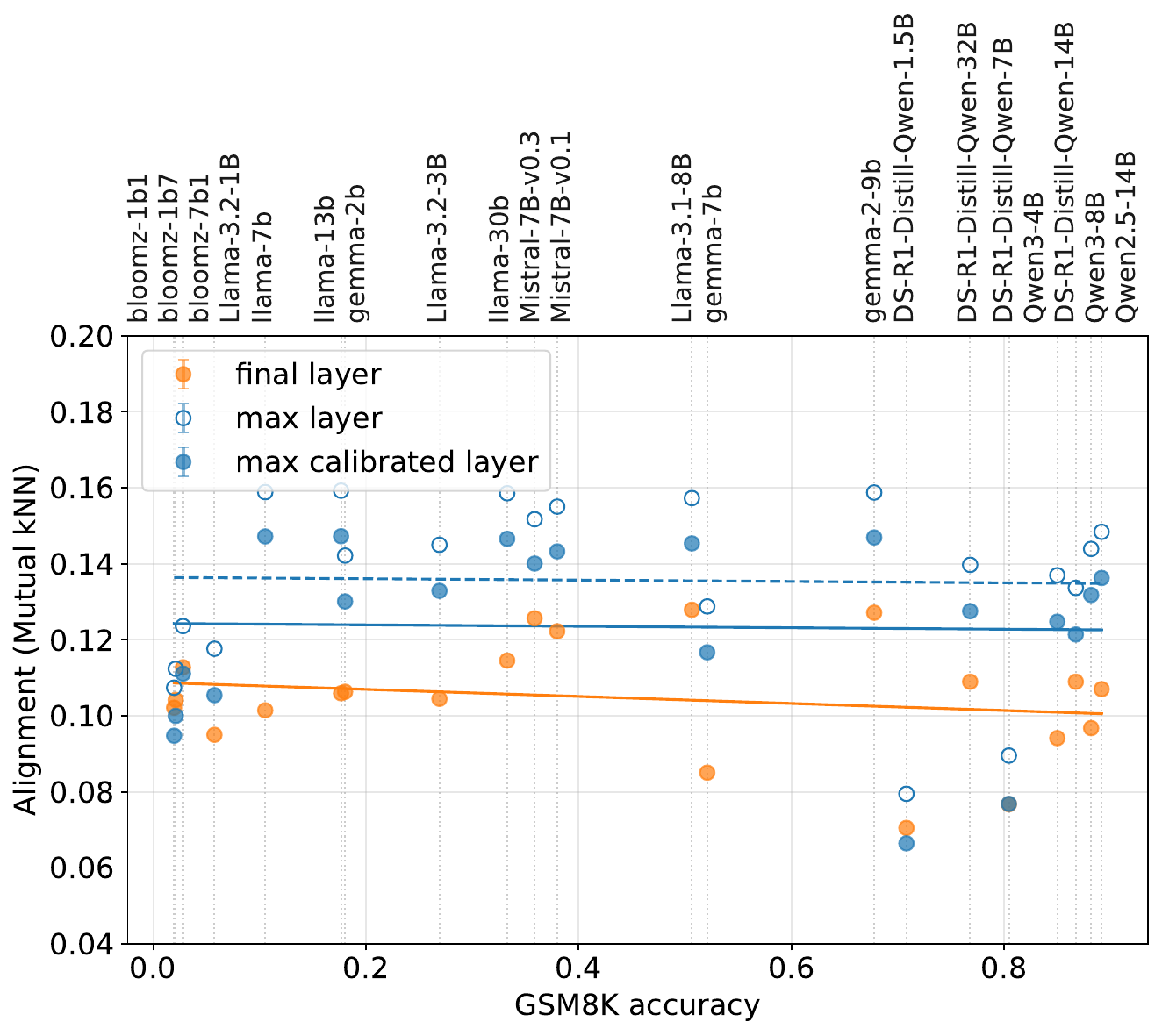}
\caption{WIT.}
\label{fig:gsm8k_alignment_wit_app}
\end{subfigure}
\hfill
\begin{subfigure}{0.48\linewidth}
\centering
\includegraphics[width=\linewidth]{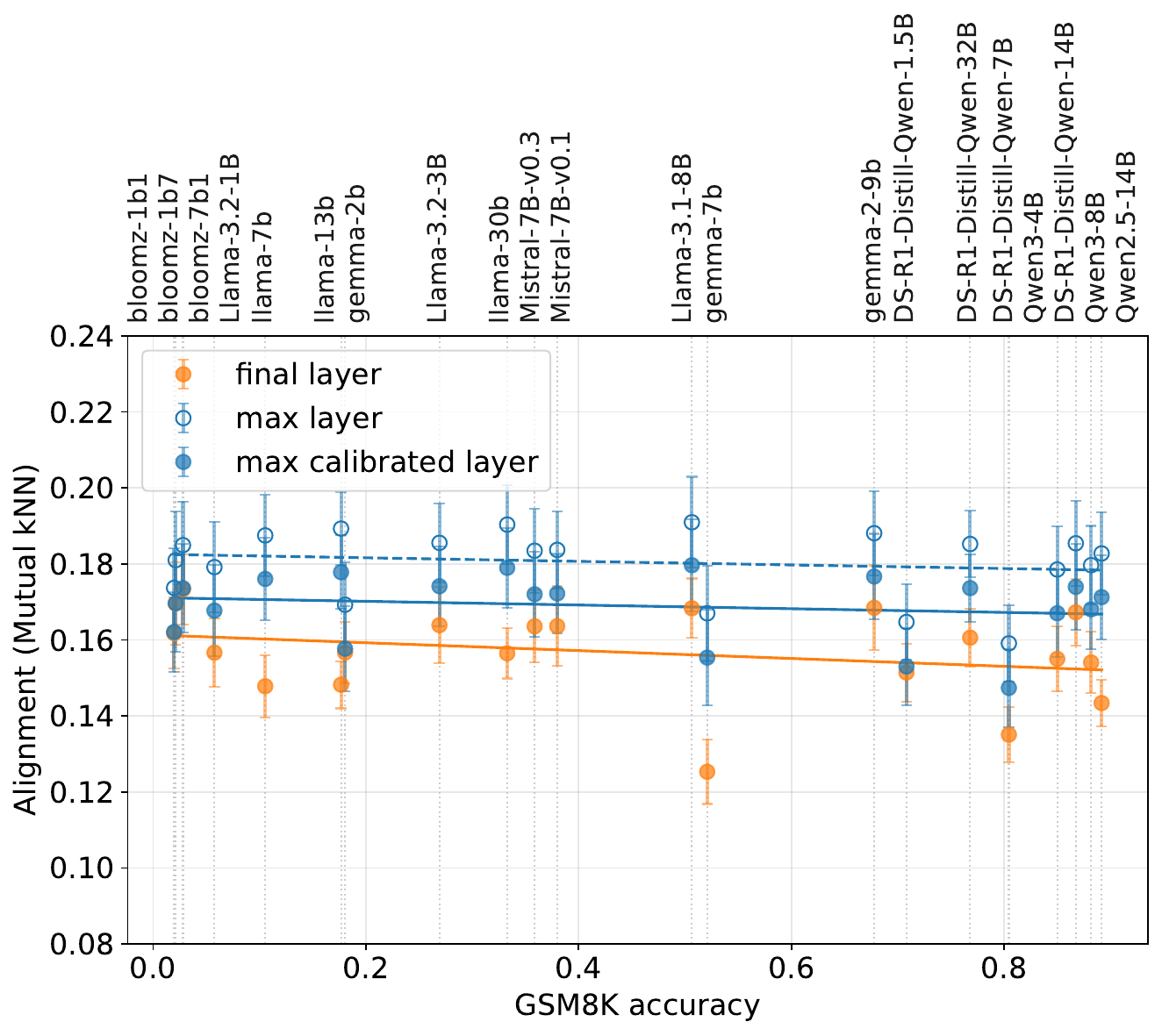}
\caption{CC3M short synthetic.}
\label{fig:gsm8k_alignment_cc3m_short_app}
\end{subfigure}
\begin{subfigure}{0.48\linewidth}
\centering
\includegraphics[width=\linewidth]{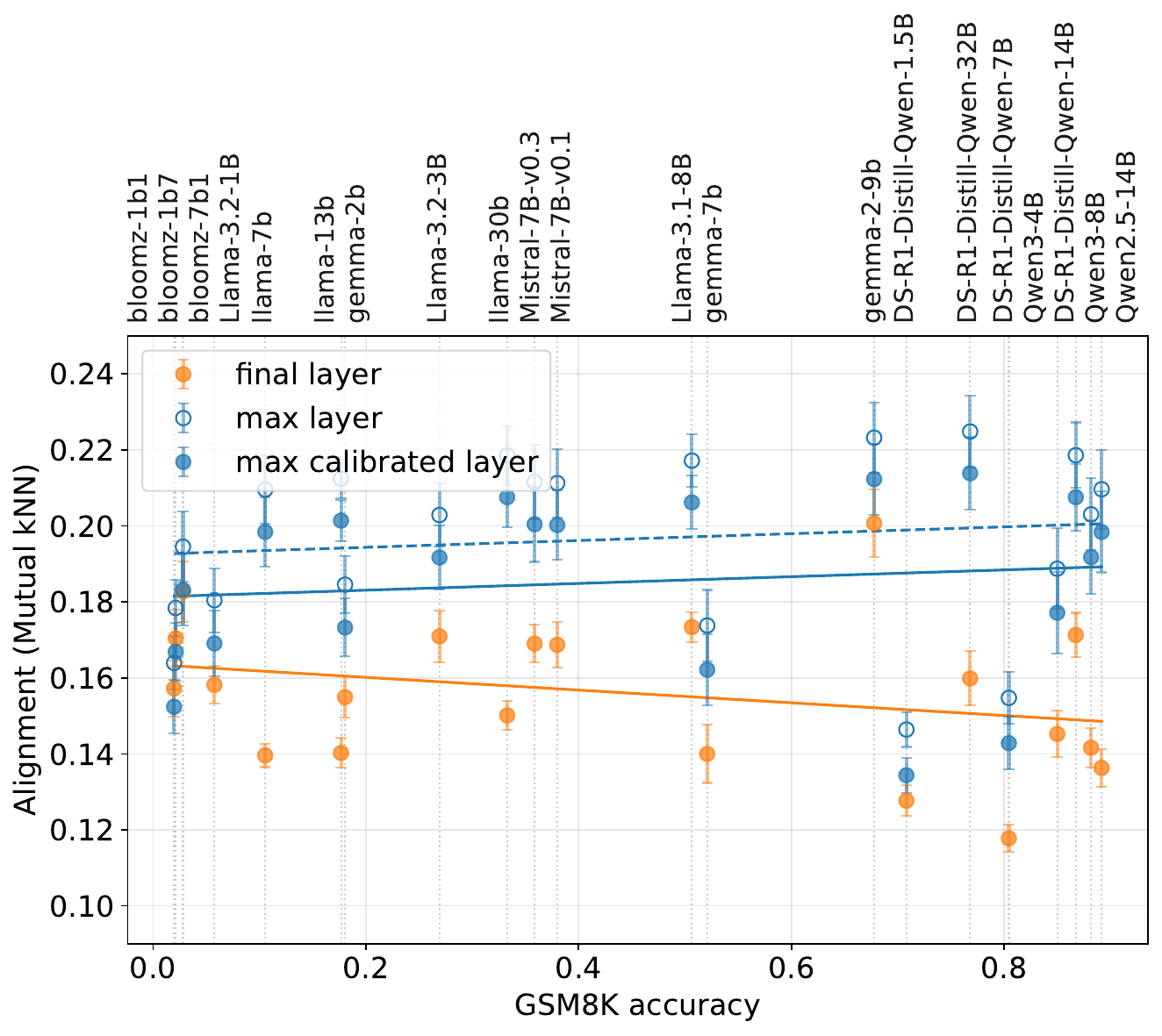}
\caption{DOCCI short synthetic.}
\label{fig:gsm8k_alignment_docci_short_app}
\end{subfigure}
\hfill
\begin{subfigure}{0.48\linewidth}
\centering
\includegraphics[width=\linewidth]{figures/mutual_knn_gsm8k_with_calibration_final/dinov3/gsm8k_mutual_knn_docci_with_summaries_description.pdf}
\caption{DOCCI.}
\label{fig:gsm8k_alignment_docci_long_app}
\end{subfigure}
\caption{Performance on GSM8K vs.\ alignment with DINOv3.}
\label{fig:gsm8k_alignment_four_datasets}
\end{figure*}

Figure~\ref{fig:gsm8k_alignment_four_datasets} presents the corresponding results for GSM8K. Unlike MMLU and ARC, GSM8K accuracy exhibits no consistent positive correlation with cross-modal alignment. Therefore, cross-modal alignment should not be interpreted as a general measure of language-model capability. Instead, its relationship with performance is benchmark-dependent and may primarily reflect capabilities associated with semantic structure shared across vision and language, which mathematical reasoning on GSM8K probes less directly.

\clearpage
\section{Layer selection for VLM baselines}
\label{apd:layer_selection_vlm_app}

\begin{figure*}[ht]
\centering

\begin{subfigure}{0.32\textwidth}
    \centering
    \includegraphics[width=\linewidth]{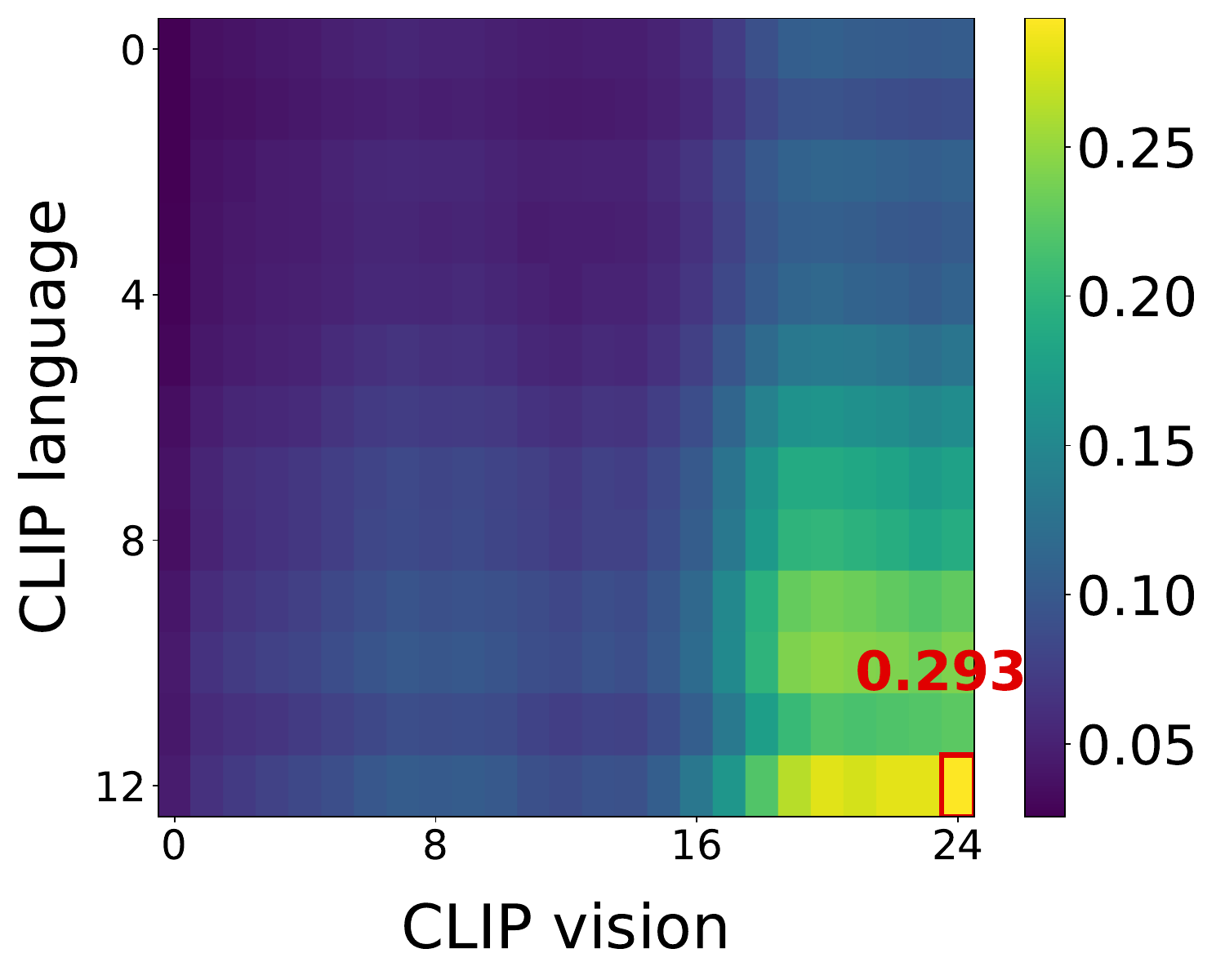}
\end{subfigure}
\hfill
\begin{subfigure}{0.32\textwidth}
    \centering
    \includegraphics[width=\linewidth]{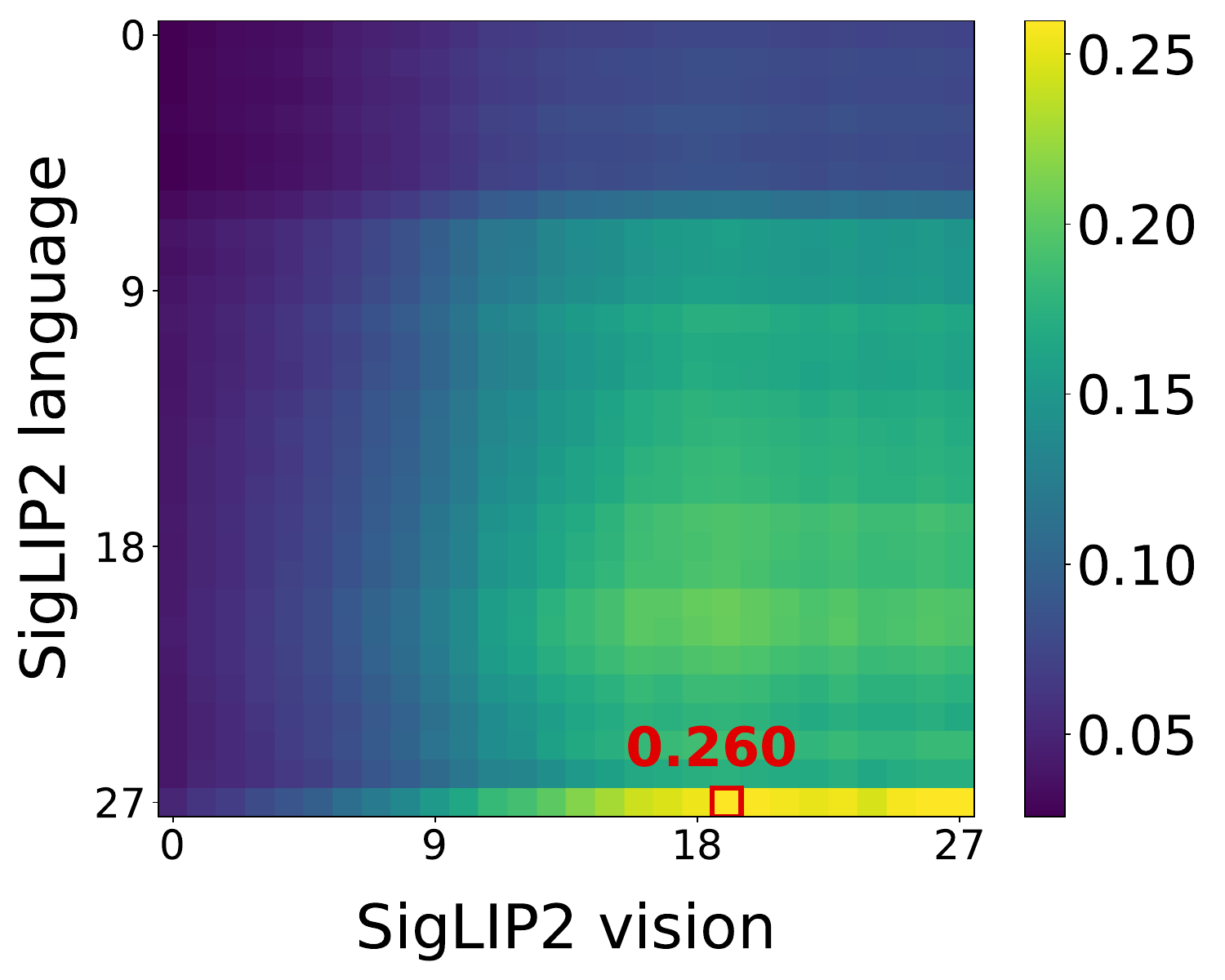}
\end{subfigure}
\hfill
\begin{subfigure}{0.32\textwidth}
    \centering
    \includegraphics[width=\linewidth]{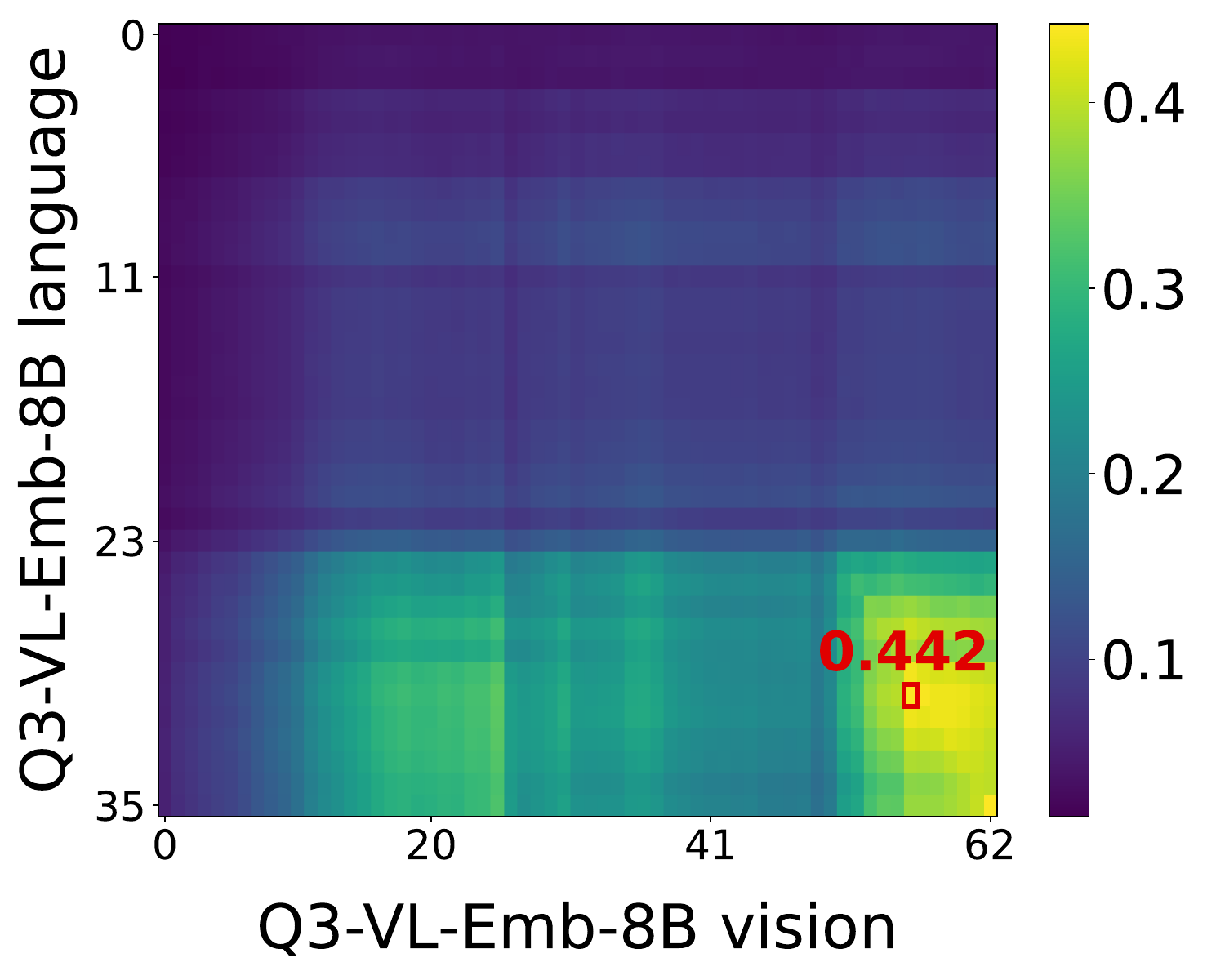}
\end{subfigure}

\caption*{VLMs}

\caption{Heatmaps with mutual $k$-NN alignment between every layer on DOCCI.}
\label{fig:heatmap_alignment_layers_vlms}
\end{figure*}

In Section~\ref{subsec:contextualization}, we represent explicitly aligned VLMs using their final layer representations in the shared image-text space. These are the representations employed in downstream tasks, such as retrieval, and directly optimized for cross-modal alignment during training. In contrast, unimodal model pairs are evaluated at their maximally aligned layer pair. To verify that this difference in evaluation does not disadvantage the VLMs, Table~\ref{tab:vlm_layers} compares the alignment obtained at the projection layer with the maximum alignment across all layer pairs. While the projection layer does not always achieve the highest score, its alignment remains consistently close to the maximum. The largest difference is 0.016 for SigLIP~2 on COCO, while all other differences are at most 0.006. Consequently, using the maximum alignment instead of the projection-layer alignment does not alter the relative ordering of the regimes in Figure~\ref{fig:alignment_scores_sorted}. The layer-wise alignment patterns of the VLMs are visualized in Figure~\ref{fig:heatmap_alignment_layers_vlms}.

\begin{table}[ht]
\caption{Alignment of explicitly aligned VLMs at the final layer pair (shared embedding space) and at the maximally aligned layer pair. Mean and standard deviation over three
runs for COCO and DOCCI. WIT consists of a single run.}
\label{tab:vlm_layers}
\begin{tabular}{lcccccc}
\toprule
 & \multicolumn{2}{c}{\textbf{WIT}} & \multicolumn{2}{c}{\textbf{COCO}} & \multicolumn{2}{c}{\textbf{DOCCI 100\%}} \\
\cmidrule(lr){2-3}\cmidrule(lr){4-5}\cmidrule(lr){6-7}
\textbf{VLM} & Final & Max & Final & Max & Final & Max \\
\midrule
CLIP & 0.199 & 0.199 & $0.317 \pm 0.005$ & $0.317 \pm 0.005$ & $0.292 \pm 0.005$ & $0.293 \pm 0.003$ \\
SigLIP 2 & 0.079 & 0.079 & $0.285 \pm 0.004$ & $0.301 \pm 0.005$ & $0.258 \pm 0.001$ & $0.263 \pm 0.006$ \\
Q3-VL & 0.218 & 0.224 & $0.407 \pm 0.002$ & $0.407 \pm 0.002$ & $0.445 \pm 0.003$ & $0.445 \pm 0.003$ \\
\bottomrule
\end{tabular}
\end{table}

\clearpage
\section{Contextualization with varying DOCCI captions}
\label{apd:contextualization_app}

\begin{figure}[ht]
\centering
\centering
\includegraphics[width=1.0\columnwidth]{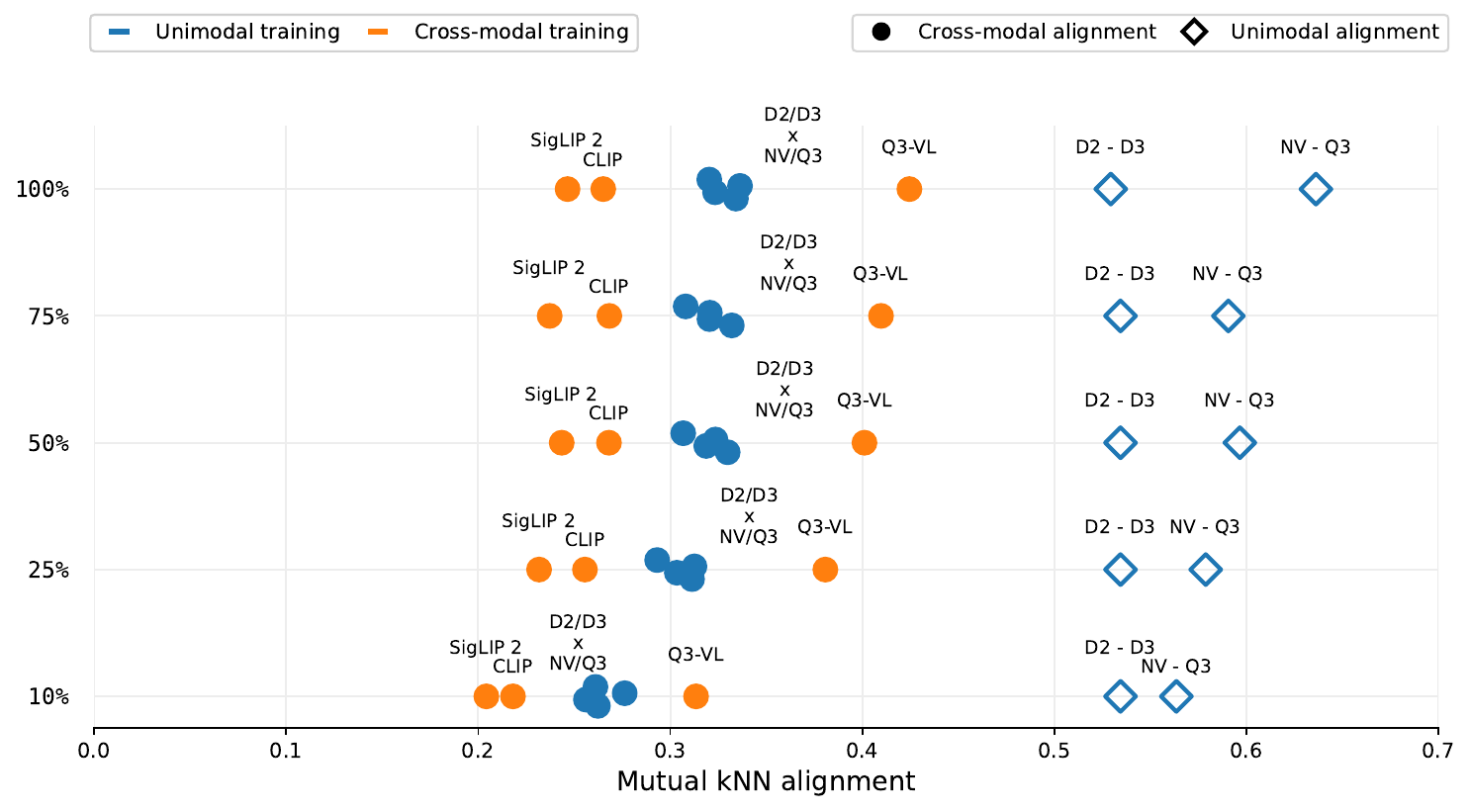}
\vspace{-0.4cm}
\caption{Alignment for different models and evaluation protocols with DOCCI captions of varying verbosity: DINOv2/v3 (D2/D3), Qwen3-Emb-8B (Q3), NV-Embed-v2 (NV), CLIP, SigLIP 2, Qwen3-VL-Emb-8B (Q3-VL).}
\label{fig:alignment_scores_app}
\end{figure}

Section~\ref{subsec:contextualization} compares the alignment of unimodal models against a set of baselines on original DOCCI captions and selected summarized variants. Figure~\ref{fig:alignment_scores_app} complements this analysis by comparing the contextualization baselines across a larger set of DOCCI caption lengths while keeping the underlying images fixed. As captions become more concise, the absolute cross-modal alignment measured by mutual $k$-NN decreases. However, the relative ordering between unimodal vision-language pairs, explicitly aligned VLMs, and within-modality pairs remains largely unchanged. In particular, DINOv2/v3 paired with Qwen3-Embedding-8B or NV-Embed-v2 remain more strongly aligned than CLIP and SigLIP~2 across caption lengths.

\clearpage
\section{Mutual $k$-NN for varying $k$ and $N$}
\label{app:mutualkNN}

We investigate the sensitivity of mutual $k$-NN alignment to the choice of the neighborhood size $k$ and the gallery size $N$ using the DOCCI dataset. 

\begin{figure*}[ht]
\centering
\includegraphics[width=1.0\columnwidth]{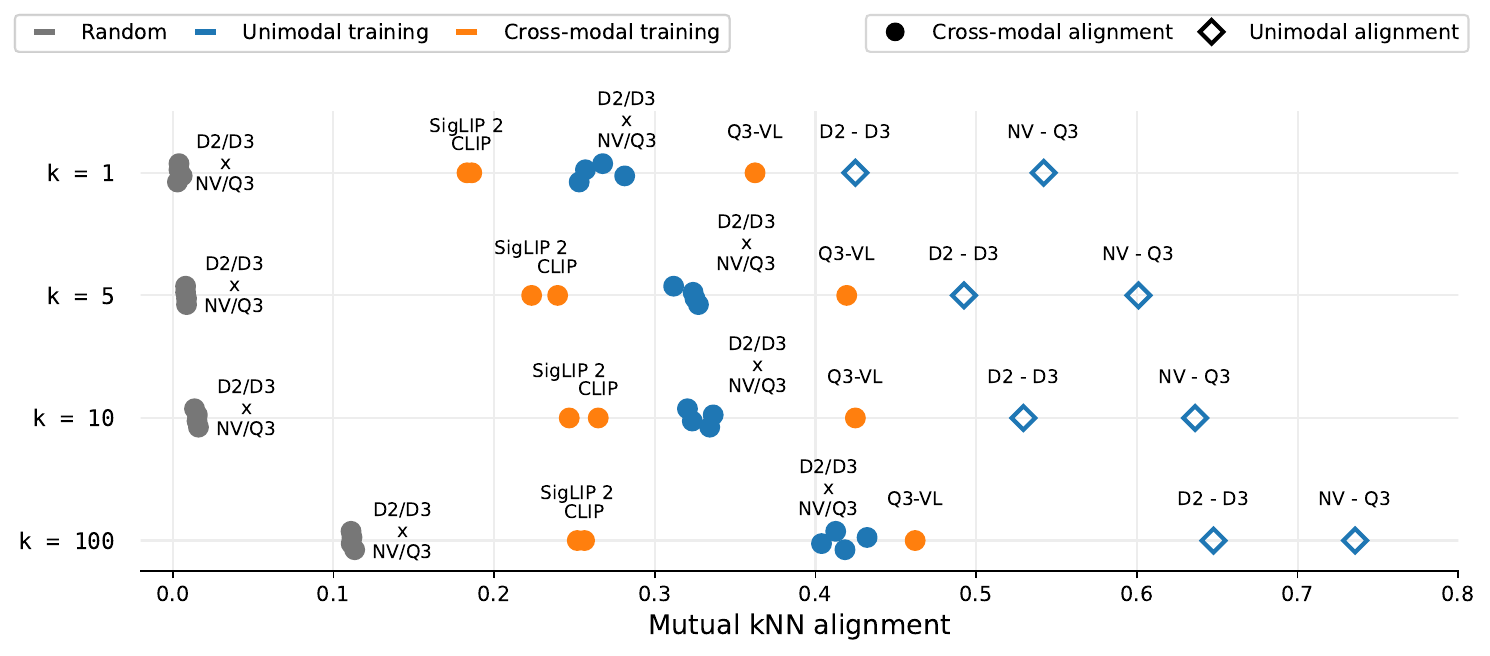}
\label{fig:alignment_scores_k_ablation}
\vspace{-0.4cm}
\caption{Alignment for different models and evaluation protocols with $N = 1024$ DOCCI samples and varying $k$.}
\label{fig:alignment_scores_sorted_k_ablation}
\end{figure*}

First, we fix the gallery size to $N = 1024$ and compare $k \in \{1, 5, 10, 100 \}$ in Figure~\ref{fig:alignment_scores_sorted_k_ablation}. While the alignment scores increase with larger $k$, the relative ordering of the alignment regimes remains unchanged, suggesting that the observed differences are robust to the choice of $k$.

\begin{figure*}[ht]
\centering
\includegraphics[width=1.0\columnwidth]{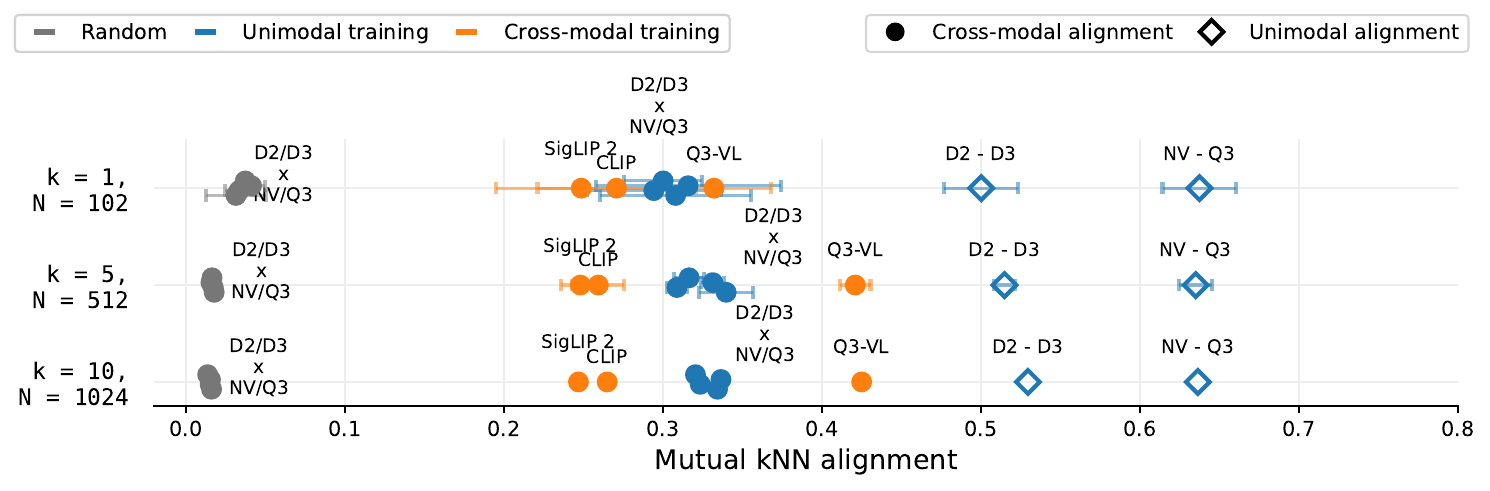}
\label{fig:alignment_scores_ratio_ablation}
\vspace{-0.4cm}
\caption{Alignment for different models and evaluation protocols on DOCCI samples with varying $N$ and $k$ but constant ratio $\frac{k}{N}$.}
\label{fig:alignment_scores_sorted_ratio_ablation}
\end{figure*}

Second, we examine the effect of varying $k$ while keeping the ratio $\frac{k}{N} \approx 0.01$ fixed, considering $(k, N) \in \{(1,102),(5,512),(10,1024)\}$ (see Figure~\ref{fig:alignment_scores_sorted_ratio_ablation}). For each gallery size ($N < 1024$), we randomly subsample from the same pool of $1024$ images and repeat the experiment five times, reporting the mean and standard deviation across subsamples. For $N = 1024$, we use the full set of images and report a single score. The alignment scores remain relatively stable across these settings, which is in line with \citet{koepke2026back}, who find that alignment is stable when the neighborhood size scales with the gallery size.

\clearpage
\section{DOCCI summarization prompt and example}
\label{apd:prompt}

\begin{figure}[ht]
    \centering
    \begin{tcolorbox}[
        colback=white,
        colframe=gray!50,
        boxrule=0.7pt,
        arc=2pt,
        left=6pt,
        right=6pt,
        top=6pt,
        bottom=6pt,
    ]

    \begin{minipage}[t]{0.32\linewidth}
        \vspace{0pt}
        \centering
        \includegraphics[width=\linewidth]{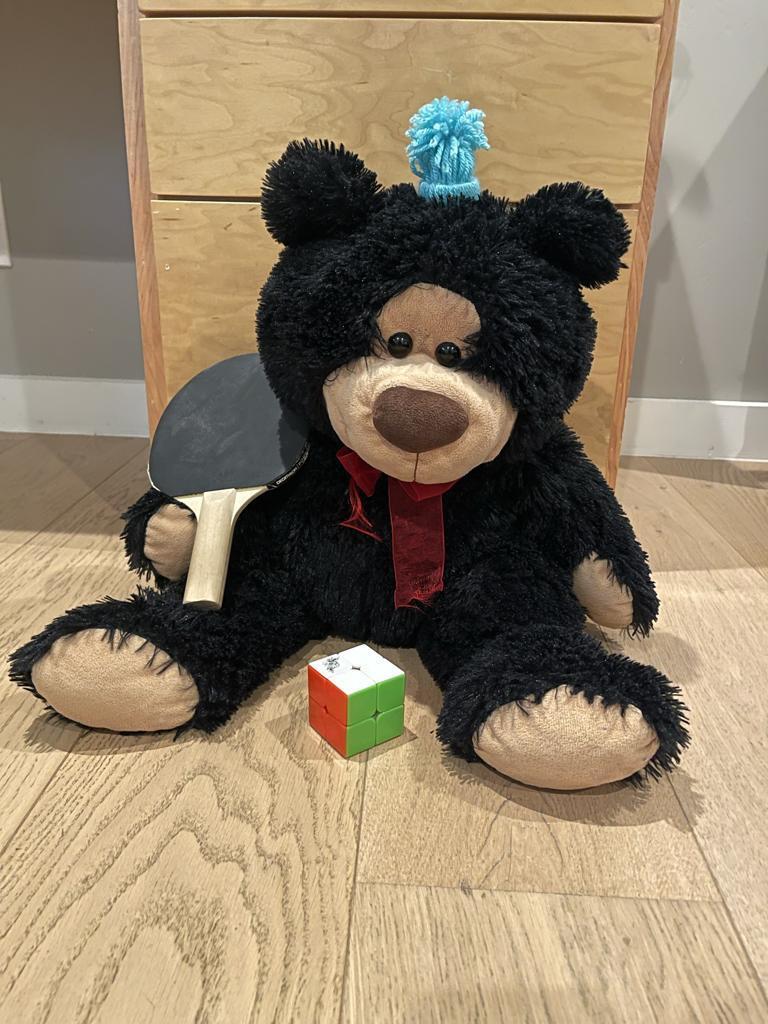}
    \end{minipage}
    \hfill
    \begin{minipage}[t]{0.64\linewidth}
        \vspace{0pt}

        \begin{tcolorbox}[
            colback=gray!15,
            colframe=gray!50,
            boxrule=0.3pt,
            arc=2pt,
            left=5pt,
            right=5pt,
            top=4pt,
            bottom=4pt,
        ]
        \textbf{\textit{Original caption}}\\[-0.2em]
        \footnotesize
        \newline
        A front view of a black stuffed teddy bear that is sitting on a light wood floor. It is propped up against a wood dresser with drawers. It has a tiny blue winter hat sitting on the top of its head. Not only that, but it has a red, tattered bow around its neck. Its paws on its feet and hands are tan. It has a dark brown, round nose and two small black eyes that are surrounded by tan fur. A ping pong paddle is sitting on it on the left by its arm. In front of it, between its legs, is a red, white and green square plastic cube game. A gray wall is behind the dresser, with white molding near the wood floor. A shadow is on the floor behind the dresser and the wall on the left. Light is shining on the right side of the wall, and there is a shadow on part of it on the far right. Light is shining on the bear and the floor in front of it.
        \end{tcolorbox}

    \end{minipage}

    \vspace{0.8em}

    \footnotesize
    \textbf{75\%}\quad
    \footnotesize
    A black stuffed teddy bear sits on a light wood floor, leaning against a wooden dresser. It wears a tiny blue winter hat and a red, tattered bow. Its paws and hands are tan, with a dark brown nose and black eyes surrounded by tan fur. A ping pong paddle rests on the left of its arm. Between its legs lies a red, white, and green plastic cube. A gray wall with white molding is behind the dresser, casting a shadow on the floor. Light shines on the right side of the wall, creating a shadow on the far right. Light also illuminates the bear and the floor in front of it.

    \vspace{0.5em}

    \textbf{50\%}\quad
    \footnotesize
    A black stuffed teddy bear sits on a wooden floor, leaning against a dresser. It wears a blue winter hat and a red, tattered bow. Its tan paws and hands rest on the floor. A ping pong paddle lies beside it. A red, white, and green cube game sits between its legs. A gray wall with white molding is behind the dresser. Shadows fall on the floor and wall, with light coming from the right.

    \vspace{0.5em}

    \textbf{25\%}\quad
    \footnotesize
    A black teddy bear sits on a wooden floor, leaning against a dresser. It wears a blue hat and a red bow. A ping pong paddle rests near its arm. A colored cube game lies between its legs. A gray wall with white molding is behind the dresser. Light shines on the bear and floor.

    \vspace{0.5em}

    \textbf{10\%}\quad
    \footnotesize
    A black teddy bear sits on a wooden floor, propped against a dresser. It wears a blue hat and red bow. A paddle and cube game are nearby. Light shines on the bear and floor.

    \end{tcolorbox}
    \caption{Example of progressively compressed DOCCI captions for the same image.}
    \label{fig:caption-compression-example}
\end{figure}

We provide the prompt for summarizing DOCCI captions to a fraction of their original word count in Figure \ref{fig:docci-summarization-prompt}. An example from DOCCI with its original caption and newly generated captions of varying length is displayed in Figure \ref{fig:caption-compression-example}.

\begin{figure}[ht]
    \centering

\begin{promptbox}
You are compressing a detailed image description to {target_ratio} of its original length.

Your goal is to reduce the LEVEL OF VISUAL DETAIL in the description.
At shorter target lengths, the output should describe substantially fewer
visual details, not merely express the same details using fewer words.
Your task is to rewrite the description as a shorter caption of {target_words} words while preserving
the important visually grounded information. The {target_words} target is a strict length requirement, not a suggestion!

IMPORTANT RULES:
- Use ONLY information contained in the original description.
- Do NOT add, infer, or invent information.
- Keep the remaining text grammatically complete and natural.
- Do NOT create sentence fragments.
- Do NOT remove function words merely to save space.
- At shorter target lengths, it is EXPECTED that many secondary details
will be completely omitted.
- Prioritize the most salient subjects and their primary actions.
- The target length is {target_words} words.
- Your caption HAS TO be LONGER than {target_words - 10} and SHORTER than {target_words + 10}.
- Write a natural, self-contained image caption.
- A simple caption containing only the main subject and action is preferable
to a dense caption containing many secondary details.
- Output ONLY the compressed caption. Do not explain your changes.

The original description contains approximately {original_words} words.

Original description:
{caption}

Compressed caption (~{target_words} words):
\end{promptbox}
\caption{Prompt for the summarization of detailed DOCCI captions with Qwen3-8B to a fraction of their original word count.}
\label{fig:docci-summarization-prompt}
\end{figure}

\end{document}